\documentclass{article}

\usepackage{microtype}
\usepackage{graphicx}
\usepackage{subcaption}
\usepackage{booktabs} 
\usepackage{multirow}
\usepackage{multicol}
\usepackage[table]{xcolor}
\usepackage{float} 

\definecolor{posgreen}{RGB}{220,245,220}
\definecolor{negred}{RGB}{245,220,220}
\definecolor{bestgreen}{RGB}{180,235,180}
\definecolor{worstred}{RGB}{235,180,180}
\definecolor{vdrow}{RGB}{230,240,255}      
\definecolor{grouprow}{RGB}{245,245,245}   
\definecolor{whitebg}{RGB}{255,255,255}  

\usepackage{hyperref}

\usepackage[preprint]{icml2026}

\usepackage{amsmath}
\usepackage{amssymb}
\usepackage{mathtools}
\usepackage{amsthm}

\usepackage[capitalize,noabbrev]{cleveref}

\theoremstyle{plain}

\theoremstyle{definition}

\theoremstyle{remark}

\usepackage[textsize=tiny]{todonotes}

\icmltitlerunning{Learning from Visual Deltas}

\begin{document}

\twocolumn[
    \icmltitle{VisualDeltas: Learning Preferences \\ from Visual Quality Perturbations}

    \begin{center}
        \textbf{Hailiang Huang}$^{1}$, 
        \textbf{Yihao Liu}$^{1}$, 
        \textbf{Shengyue Guan}$^{2}$, 
        \textbf{Haoze Li}$^{3}$, 
        \textbf{Sujian Li}$^{1\dagger}$ \\
        \vspace{5pt}
        $^{1}$Peking University, Beijing, China \\
        $^{2}$Alibaba Group, Hangzhou, China \\
        $^{3}$Zhejiang University, Hangzhou, China \\
        \vspace{5pt}
        {\small \texttt{\{hailiang, haoeliu\}@stu.pku.edu.cn, lisujian@pku.edu.cn}}
    \end{center}

    \icmlcorrespondingauthor{Sujian Li$^{\dagger}$}{lisujian@pku.edu.cn}
    
    \vskip 0.3in
]



\printAffiliationsAndNotice{}  

\begin{abstract}
We present VisualDeltas, a lightweight preference-learning framework that extracts supervision from visual quality variations in multimodal data. By leveraging the systematic impact of image quality on visual perception and reasoning, VisualDeltas induces informative preference signals without relying on human annotations or external teachers. The framework supports both label-free and label-based regimes, enabling flexible use of available supervision when present. Across diverse multimodal benchmarks and model scales, VisualDeltas consistently outperforms rejection-sampling fine-tuning and improves generalization, and extends naturally to a range of visual degradations.
\end{abstract}

\section{Introduction}
\label{sec:introduction}

\begin{figure}[t]
    \centering
    \includegraphics[width=\linewidth]{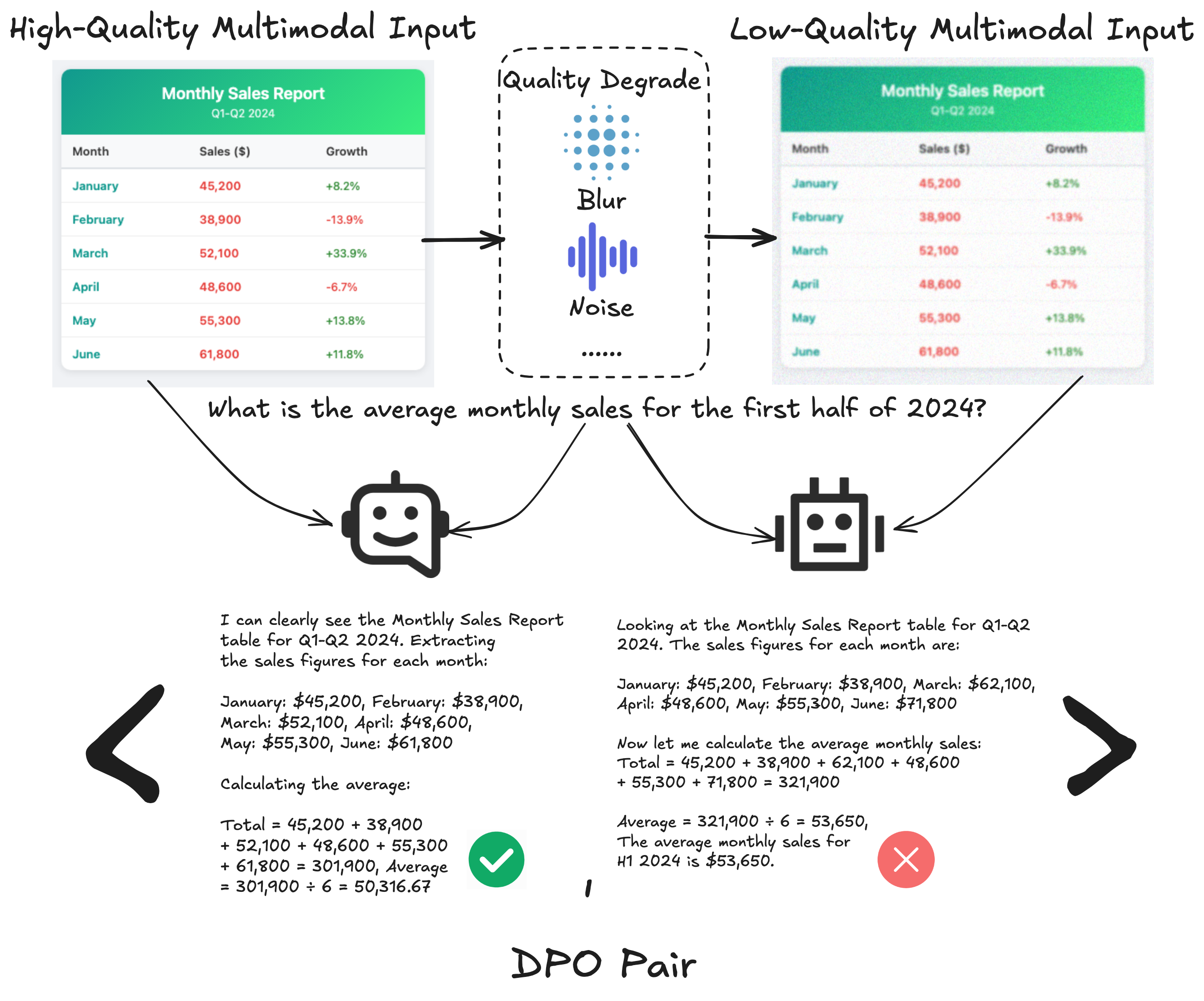}
    \caption{
    DPO pair construction via multimodal input quality perturbation.
    HQ and LQ inputs for the same multimodal QA task induce correct and incorrect reasoning, forming a natural preference pair.
    }
    \label{fig:dpo_delta_multimodal}
\end{figure}

Modern vision-language models have made rapid progress on multimodal question answering over images, documents, charts, and tables, yet improving their reasoning ability still often depends on costly supervision pipelines---large-scale labeled datasets, external preference annotation, or RLHF-style optimization with reward models and judges \citep{rafailov2023direct,bai2022constitutional,yuan2024self}. This creates a practical bottleneck: for many multimodal tasks, we would like a lightweight post-training recipe that improves the model without introducing new annotators, reward models, or stronger teacher systems.

A key but underexploited property of multimodal QA systems is their intrinsic sensitivity to visual input quality. Even when the underlying instance and question remain unchanged, controlled degradations such as resolution reduction can impair perceptual clarity (e.g., text legibility or structural alignment) and lead to inconsistent reasoning trajectories and unstable predictions \citep{li2025vision,zhou2025robust,fan2025v,tang2025robust}. Crucially, this sensitivity does not only reveal brittleness---it naturally produces paired model behaviors on the same QA instance: the model answers once under an HQ view and once under an LQ view, often yielding a meaningful relative quality gap between the two responses.

This paper proposes a simple perspective: rather than treating perturbations primarily as a robustness evaluation tool or as an input-side training trick, we use controlled visual quality changes as a mechanism to generate relative supervision. Concretely, we query the same vision-language model on a HQ input and a LQ counterpart, and treat the resulting responses as a preference pair for direct preference optimization (Fig.~\ref{fig:dpo_delta_multimodal}). This turns the model's own quality sensitivity into scalable preference data---without external judges, reward models, or stronger models \citep{geng2025delta,saeidi2024triple}.

Our approach is inspired by the broader idea that relative deltas can be sufficient for preference learning even when absolute supervision is imperfect: paired comparisons can provide a direction of improvement that supervised finetuning on weak targets may not \citep{yin2024relative,chen2024mallowspo}. We instantiate this ``delta supervision'' principle in the multimodal setting by constructing preference pairs induced purely by input-quality variation. Notably, we observe that degraded visual inputs often trigger compensatory but ineffective reasoning behaviors---such as generating longer yet less accurate responses---which provide natural negative samples for preference learning. Because the $\text{HQ}\succ\text{LQ}$ heuristic can be noisy for some instances, we also analyze when it is reliable and study simple selection strategies (including a label-assisted variant) that retain the most informative pairs while preserving the lightweight nature of the pipeline.

We call the resulting framework VisualDeltas. Across multiple multimodal QA benchmarks and model scales, VisualDeltas consistently improves answer accuracy and generalization compared to correctness-only fine-tuning baselines. While we focus on resolution as the simplest controllable knob, the same principle extends to other semantics-preserving degradations: Gaussian noise and motion blur also induce effective preference signals and yield comparable gains.

\paragraph{Contributions.}
We make three main contributions:
\begin{itemize}
    \item We introduce \textbf{VisualDeltas}, a preference learning framework that exploits resolution-induced response deltas to construct preference pairs without requiring external annotation or reward models.
    \item We show that simple, controllable visual degradations (e.g., resolution reduction) consistently elicit informative response deltas that can be exploited as preference supervision.
    \item We validate the approach on multiple benchmarks, demonstrating consistent gains over correctness-only fine-tuning.
\end{itemize}

\section{Related Work}
\paragraph{Visual Perturbations for Multimodal Reasoning.}
Recent studies show that multimodal reasoning models are highly sensitive to visual input quality, where even mild perturbations can substantially change intermediate reasoning behavior and final predictions~\citep{fan2025v}. Fan et al.~\citep{fan2025v} introduced V$^2$R-Bench to holistically evaluate the robustness of large vision--language models (LVLMs) under fundamental visual variations, highlighting that such sensitivity is widespread across mainstream models. Beyond benchmarking, prior work has investigated how degradations such as blur, occlusion, and compression affect multimodal reasoning: Li et al.~\citep{li2025vision} demonstrated that simple perturbations can significantly modulate multimodal mathematical reasoning performance, while Tang et al.~\citep{tang2025robust} proposed degradation-aware reasoning mechanisms to mitigate failures under degraded inputs. Zhou et al.~\citep{zhou2025robust} further analyzed scientific diagrams and showed that different perturbation types expose brittle visual understanding, motivating robustness-oriented training strategies.

Most existing studies treat perturbations primarily as a means to \emph{evaluate} robustness or to \emph{augment} training for better average performance~\citep{zhou2025robust,li2025vision}. Our perspective is different: we use controlled perturbations as a mechanism to \emph{generate relative supervision}. Concretely, we control image/table resolution to keep the underlying instance and QA target fixed while modulating perceptual quality (e.g., legibility and structural clarity). This often induces divergent model responses for the same multimodal QA input, enabling the construction of preference pairs without external judges. In this sense, our method differs from perturbation-based training approaches that primarily optimize robustness using labeled supervision or augmentation pipelines (e.g., perturbation-driven reliability optimization or distractor-based robustness training)~\citep{gumutualvpr,li2025vision}.

\paragraph{DPO and Preference Data Construction.}
DPO provides an efficient alternative to RLHF by optimizing directly on paired preferences without training an explicit reward model~\citep{rafailov2023direct}. Beyond human-labeled comparisons, recent work explores self-generated or synthetic preferences, for example by degrading high-quality outputs, sampling multiple candidate responses, or synthesizing preference labels with auxiliary critics or models~\citep{bai2022constitutional,yuan2024self,cui2023ultrafeedback}.

Many such pipelines require additional components (e.g., external teachers/critics, generators, or extensive post-hoc scoring and filtering), which can increase complexity and introduce bias~\citep{yuan2024self,cui2023ultrafeedback}. In contrast, our preference pairs are constructed from the \emph{same} model's responses under controlled resolution changes for the same QA input, requiring no additional human annotation and no auxiliary teacher/critic model. The preference signal arises from response-level deltas induced by input-quality variation, and we empirically characterize when this heuristic direction is reliable.

\paragraph{Preference Optimization in Multimodal Alignment.}
Preference optimization has been extended to multimodal alignment and reasoning, with methods addressing issues such as modality overfitting, difficulty imbalance, and reasoning alignment~\citep{jiang2024modality,gao2025m3po,lu2025adavip,wang2024mdpo,zhang2025mm,wu2025mitigating,zhang2025videoDPO,pesaran2025lpoi}. Many existing approaches obtain preferences via external supervision (human labels or model-based scoring), retrieval-augmented pairing, or explicit editing/constraint of intermediate reasoning traces, which can incur additional cost and introduce extra design choices~\citep{jiang2024modality,gao2025m3po,wang2024mdpo,wu2025mitigating}.

Our approach differs primarily in preference \emph{construction}: rather than relying on external preference sources or edited reasoning paths, we derive preferences intrinsically from the model's own sensitivity to multimodal input quality. This makes preference generation lightweight and scalable, and ties the supervision signal directly to failure modes triggered by degraded perception.

\paragraph{Delta Learning and Relative Supervision.}
The delta learning paradigm argues that relative differences between model outputs can serve as effective supervision, especially when absolute labels are noisy or weak~\citep{geng2025delta,yin2024relative,chen2024mallowspo}. While relative supervision has been widely explored in text generation and dialogue via contrastive preferences or paired ranking objectives~\citep{yin2024relative,chen2024mallowspo}, its use as a \emph{systematic} supervision source for multimodal reasoning remains less explored.

Some prior work leverages differential visual signals in robustness-oriented training or in multimodal preference optimization variants, for example by incorporating perturbations to improve robustness~\citep{li2025vision} or by adding multimodal-specific constraints in DPO-style objectives~\citep{wang2024mdpo}. In contrast, we present a simple instantiation of delta learning for multimodal QA where controlled resolution changes are used to produce paired responses for the same input, and we integrate these pairs with DPO for stable preference optimization.

\paragraph{Summary.}
In contrast to prior work that studies visual perturbations, preference learning, or multimodal alignment in isolation, our work unifies these directions. We show that resolution-induced reasoning deltas provide a lightweight, intrinsic source of preference supervision for multimodal question answering, enabling scalable and effective preference optimization without external annotation or reward models.

\section{Methodology}
\label{sec:method}

\begin{figure*}[!t]
    \centering
    \includegraphics[width=0.85\linewidth]{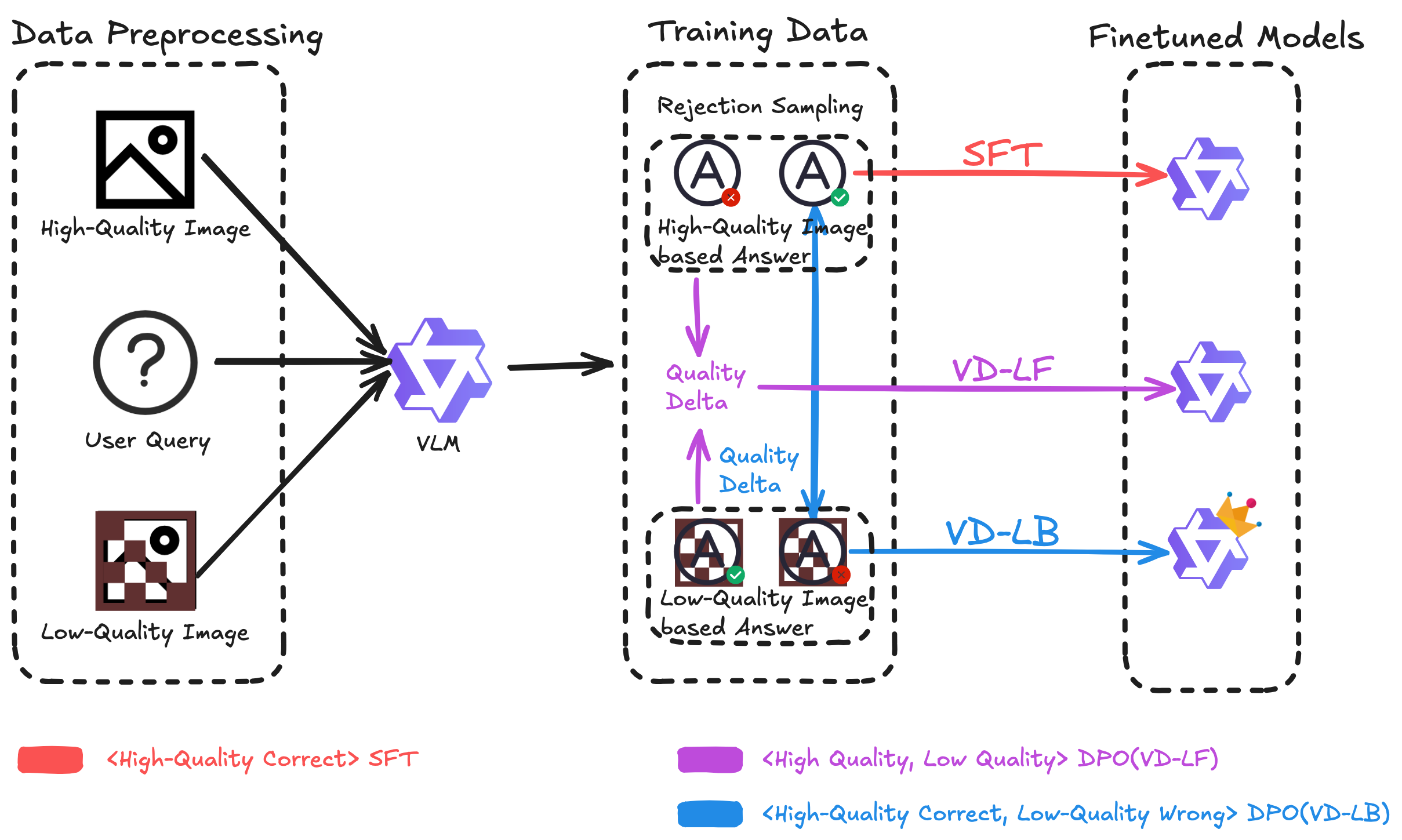}
    \caption{
    Overview of VisualDeltas. For each user query, the model generates paired responses under both HQ and LQ image views. We compare with \textbf{SFT} (supervised fine-tuning on HQ-correct responses only) and two VisualDeltas variants: \textbf{VD-LF} (label-free) uses \emph{all} HQ vs.\ LQ pairs without correctness filtering, while \textbf{VD-LB} (label-based) selects only HQ-correct vs.\ LQ-wrong pairs. Both VisualDeltas variants apply DPO exclusively with HQ context during training.
    }
    \label{fig:method_overview}
\end{figure*}

This section provides a formalization of VisualDeltas to clarify its foundations and implementation. Since our approach builds on DPO, we present the mathematical formulation of how visual quality variations induce preference pairs and how these pairs are used for training. This formalization makes precise the conceptual pipeline introduced in \ref{sec:introduction} and serves as the basis for the empirical study in \ref{sec:experiments}.

\subsection{Preference Pair Construction}

VisualDeltas constructs preference pairs by comparing model outputs under HQ and LQ visual inputs. For a fixed VLM, we exploit the systematic quality difference: outputs from degraded images tend to be less accurate than those from original images.

\paragraph{Data preparation.}
Given a dataset $\mathcal{D} = \{(x_i, v_i)\}_{i=1}^{N}$ of text-image pairs, we create two visual views for each sample by applying a controlled degradation operator $\mathcal{T}_\alpha$ of severity $\alpha$:
\begin{equation}
v_i^{HQ} = v_i, \qquad v_i^{LQ} = \mathcal{T}_\alpha(v_i).
\end{equation}
The HQ view preserves the original image quality, while the LQ view simulates degraded visual input. The operator $\mathcal{T}_\alpha$ is abstract and general, representing any transformation that systematically reduces image quality (e.g., resolution reduction, blur, noise, compression artifacts). For models that support variable-resolution input, $v_i^{LQ}$ can be used directly without resizing or additional preprocessing.

\paragraph{Output generation.}
Using a pretrained VLM policy $\pi_{\theta_0}$, we generate responses for both views under the same text query $x_i$:
\begin{align}
o_i^{HQ} &\sim \pi_{\theta_0}(\cdot \mid x_i, v_i^{HQ}), \label{eq:hq_lq_outputs} \\
o_i^{LQ} &\sim \pi_{\theta_0}(\cdot \mid x_i, v_i^{LQ}).
\end{align}
This yields paired outputs that typically differ in quality due to the visual degradation.

\paragraph{Preference pair construction.}
From these outputs, we form preference tuples $(c_i^{HQ}, o_i^{HQ}, o_i^{LQ})$ where $c_i^{HQ} = (x_i, v_i^{HQ})$ is the HQ context. The preference relation between the two outputs depends on whether ground-truth labels are available:

\subparagraph{Label-free.}
When no ground-truth labels exist, we use the heuristic rule $o_i^{HQ} \succ o_i^{LQ}$, assuming higher visual quality leads to better responses. All generated pairs are used for training.

\subparagraph{Label-based.}
When ground-truth answers $y_i$ are available, we enforce a stricter criterion: retain only pairs where the HQ output is correct and the LQ output is wrong. Formally, let $\text{Corr}(o, y)$ indicate whether output $o$ matches ground truth $y$. We define:
\begin{equation}
\begin{split}
\mathcal{D}_{LB} = \{(c_i^{HQ}, o_i^{HQ}, o_i^{LQ}) \mid & \text{Corr}(o_i^{HQ}, y_i) \\
& \land \neg\text{Corr}(o_i^{LQ}, y_i)\}.
\end{split}
\end{equation}
This provides cleaner supervision by ensuring the preference captures clear quality-induced reasoning failures.

\paragraph{Training-time conditioning.}
Both label-free and label-based variants use DPO training that conditions only on the HQ context $c_i^{HQ}$. The LQ image acts solely as a mechanism to generate a negative sample during pair construction, ensuring consistency between training and inference.

\subsection{HQ-Conditioned Direct Preference Optimization}

DPO training conditions only on the HQ context $c_i^{HQ}$. Let $\pi_\text{ref} = \pi_{\theta_0}$ denote a frozen reference policy. For any output $o$ under context $c$, define:
\begin{equation}
\Delta_\theta(c, o) = \log \pi_\theta(o \mid c) - \log \pi_\text{ref}(o \mid c).
\end{equation}

The VisualDeltas objective is:
\begin{equation}
\begin{split}
\mathcal{L}_{VD}(\theta) = - \mathbb{E}_i \Big[ \log \sigma \big( \beta ( & \Delta_\theta(c_i^{HQ}, o_i^{HQ}) \\
& - \Delta_\theta(c_i^{HQ}, o_i^{LQ}) ) \big) \Big],
\end{split}
\end{equation}
where $\sigma$ is the logistic sigmoid and $\beta > 0$ is the temperature hyperparameter.

\paragraph{Interpretation.}
The LQ-induced output $o_i^{LQ}$ is treated as a negative sample. The model is trained to avoid producing this weaker output when given the HQ image, maintaining training-inference consistency. Both label-free and label-based variants share the same DPO objective; only the training set construction differs as described in \ref{sec:method}.

\begin{table*}[!t]
\centering
\footnotesize
\setlength{\tabcolsep}{3pt}
\renewcommand{\arraystretch}{0.95}
\begin{tabular}{llcccccc}
\toprule
\multirow{2}{*}{Model} & \multirow{2}{*}{Method} & \multicolumn{5}{c}{Datasets} \\
\cmidrule(lr){3-7}
& & HiTab & WikiTQ & VQA & GQA & MathVision \\
\midrule
\multicolumn{7}{l}{\textbf{Qwen2.5-7B-VL}} \\
\midrule
& \textit{Inference Only (No Training)} & & & & & \\
& Baseline & 67.93 & 66.40 & 60.00 & 39.95 & 24.87 \\
\midrule
\rowcolor{grouprow}
& \textit{Train Set: HiTab} & & & & & \\
& SFT (HQ Correct Only)
& \cellcolor{posgreen}69.89 {\footnotesize(+1.96)}
& 65.90 {\footnotesize(-0.50)}
& 60.85 {\footnotesize(+0.85)}
& 39.05 {\footnotesize(-0.90)}
& \cellcolor{worstred}22.76 {\footnotesize(-2.11)} \\
\rowcolor{vdrow}
& VD-LF (HQ vs.\ LQ)
& \cellcolor{bestgreen}\textbf{71.91} {\footnotesize(+3.98)}
& \cellcolor{posgreen}67.15 {\footnotesize(+0.75)}
& \cellcolor{posgreen}62.65 {\footnotesize(+2.65)}
& \cellcolor{posgreen}42.05 {\footnotesize(+2.10)}
& \cellcolor{whitebg}25.16 {\footnotesize(+0.29)} \\
\rowcolor{vdrow}
& VD-LB (HQ Correct vs.\ LQ Wrong)
& \cellcolor{posgreen}71.40 {\footnotesize(+3.47)}
& \cellcolor{bestgreen}\textbf{69.90} {\footnotesize(+3.50)}
& \cellcolor{bestgreen}\textbf{63.55} {\footnotesize(+3.55)}
& \cellcolor{bestgreen}\textbf{42.90} {\footnotesize(+2.95)}
& \cellcolor{whitebg}25.13 {\footnotesize(+0.26)} \\
\midrule
\rowcolor{grouprow}
& \textit{Train Set: VQA} & & & & & \\
& SFT (HQ Correct Only)
& \cellcolor{worstred}65.40 {\footnotesize(-2.53)}
& \cellcolor{worstred}62.85 {\footnotesize(-3.55)}
& \cellcolor{posgreen}65.20 {\footnotesize(+5.20)}
& \cellcolor{posgreen}42.10 {\footnotesize(+2.15)}
& \cellcolor{worstred}18.36 {\footnotesize(-6.51)} \\
\rowcolor{vdrow}
& VD-LF (HQ vs.\ LQ)
& \cellcolor{posgreen}68.81 {\footnotesize(+0.88)}
& \cellcolor{whitebg}65.40 {\footnotesize(-1.00)}
& \cellcolor{posgreen}64.80 {\footnotesize(+4.80)}
& \cellcolor{posgreen}44.60 {\footnotesize(+4.65)}
& \cellcolor{whitebg}25.66 {\footnotesize(+0.79)} \\
\rowcolor{vdrow}
& VD-LB (HQ Correct vs.\ LQ Wrong)
& \cellcolor{bestgreen}\textbf{70.58} {\footnotesize(+2.65)}
& \cellcolor{posgreen}68.15 {\footnotesize(+1.75)}
& \cellcolor{bestgreen}\textbf{68.20} {\footnotesize(+8.20)}
& \cellcolor{bestgreen}\textbf{44.75} {\footnotesize(+4.80)}
& \cellcolor{whitebg}25.36 {\footnotesize(+0.49)} \\
\midrule
\rowcolor{grouprow}
& \textit{Train Set: GQA} & & & & & \\
& SFT (HQ Correct Only)
& \cellcolor{worstred}64.27 {\footnotesize(-3.66)}
& 66.05 {\footnotesize(-0.35)}
& \cellcolor{posgreen}64.85 {\footnotesize(+4.85)}
& \cellcolor{posgreen}43.05 {\footnotesize(+3.10)}
& \cellcolor{worstred}18.09 {\footnotesize(-6.78)} \\
\rowcolor{vdrow}
& VD-LF (HQ vs.\ LQ)
& \cellcolor{posgreen}70.20 {\footnotesize(+2.27)}
& \cellcolor{posgreen}68.20 {\footnotesize(+1.80)}
& \cellcolor{posgreen}63.90 {\footnotesize(+3.90)}
& \cellcolor{posgreen}43.35 {\footnotesize(+3.40)}
& \cellcolor{whitebg}25.23 {\footnotesize(+0.36)} \\
\rowcolor{vdrow}
& VD-LB (HQ Correct vs.\ LQ Wrong)
& \cellcolor{posgreen}69.19 {\footnotesize(+1.26)}
& \cellcolor{posgreen}69.60 {\footnotesize(+3.20)}
& \cellcolor{bestgreen}\textbf{66.00} {\footnotesize(+6.00)}
& \cellcolor{bestgreen}\textbf{47.50} {\footnotesize(+7.55)}
& \cellcolor{whitebg}24.54 {\footnotesize(-0.33)} \\
\midrule[\heavyrulewidth]
\multicolumn{7}{l}{\textbf{Qwen2.5-3B-VL}} \\
\midrule
& \textit{Inference Only (No Training)} & & & & & \\
& Baseline & 58.96 & 38.45 & 56.30 & 36.30 & 17.76 \\
\midrule
\rowcolor{grouprow}
& \textit{Train Set: HiTab} & & & & & \\
& SFT (HQ Correct Only)
& \cellcolor{bestgreen}\textbf{63.45} {\footnotesize(+4.48)}
& \cellcolor{posgreen}45.75 {\footnotesize(+7.30)}
& \cellcolor{posgreen}59.40 {\footnotesize(+3.10)}
& \cellcolor{bestgreen}\textbf{41.00} {\footnotesize(+4.70)}
& \cellcolor{worstred}13.45 {\footnotesize(-4.31)} \\
\rowcolor{vdrow}
& VD-LF (HQ vs.\ LQ)
& \cellcolor{posgreen}61.62 {\footnotesize(+2.65)}
& \cellcolor{bestgreen}\textbf{52.15} {\footnotesize(+13.70)}
& \cellcolor{posgreen}60.40 {\footnotesize(+4.10)}
& \cellcolor{posgreen}40.65 {\footnotesize(+4.35)}
& \cellcolor{posgreen}18.78 {\footnotesize(+1.02)} \\
\rowcolor{vdrow}
& VD-LB (HQ Correct vs.\ LQ Wrong)
& \cellcolor{posgreen}61.11 {\footnotesize(+2.15)}
& \cellcolor{posgreen}51.35 {\footnotesize(+12.90)}
& \cellcolor{bestgreen}\textbf{60.85} {\footnotesize(+4.55)}
& \cellcolor{posgreen}39.75 {\footnotesize(+3.45)}
& \cellcolor{whitebg}18.55 {\footnotesize(+0.79)} \\
\midrule
\rowcolor{grouprow}
& \textit{Train Set: VQA} & & & & & \\
& SFT (HQ Correct Only)
& \cellcolor{worstred}56.25 {\footnotesize(-2.71)}
& \cellcolor{posgreen}48.30 {\footnotesize(+9.85)}
& \cellcolor{posgreen}59.30 {\footnotesize(+3.00)}
& 37.15 {\footnotesize(+0.85)}
& \cellcolor{worstred}11.97 {\footnotesize(-5.79)} \\
\rowcolor{vdrow}
& VD-LF (HQ vs.\ LQ)
& \cellcolor{posgreen}60.23 {\footnotesize(+1.26)}
& \cellcolor{posgreen}51.45 {\footnotesize(+13.00)}
& \cellcolor{posgreen}60.05 {\footnotesize(+3.75)}
& \cellcolor{posgreen}41.30 {\footnotesize(+5.00)}
& \cellcolor{whitebg}17.40 {\footnotesize(-0.36)}\\
\rowcolor{vdrow}
& VD-LB (HQ Correct vs.\ LQ Wrong)
& \cellcolor{whitebg}59.09 {\footnotesize(+0.13)}
& \cellcolor{bestgreen}\textbf{52.45} {\footnotesize(+14.00)}
& \cellcolor{bestgreen}\textbf{61.45} {\footnotesize(+5.15)}
& \cellcolor{bestgreen}\textbf{42.50} {\footnotesize(+6.20)}
& \cellcolor{negred}16.68 {\footnotesize(-1.08)}\\
\midrule
\rowcolor{grouprow}
& \textit{Train Set: GQA} & & & & & \\
& SFT (HQ Correct Only)
& \cellcolor{worstred}55.62 {\footnotesize(-3.35)}
& \cellcolor{posgreen}48.15 {\footnotesize(+9.70)}
& 56.85 {\footnotesize(+0.55)}
& \cellcolor{posgreen}39.55 {\footnotesize(+3.25)}
& \cellcolor{worstred}10.76 {\footnotesize(-7.00)} \\
\rowcolor{vdrow}
& VD-LF (HQ vs.\ LQ)
& \cellcolor{posgreen}60.23 {\footnotesize(+1.26)}
& \cellcolor{bestgreen}\textbf{51.70} {\footnotesize(+13.25)}
& \cellcolor{bestgreen}\textbf{58.70} {\footnotesize(+2.40)}
& \cellcolor{posgreen}41.25 {\footnotesize(+4.95)}
& \cellcolor{negred}16.68 {\footnotesize(-1.08)} \\
\rowcolor{vdrow}
& VD-LB (HQ Correct vs.\ LQ Wrong)
& \cellcolor{bestgreen}\textbf{61.36} {\footnotesize(+2.40)}
& \cellcolor{posgreen}50.60 {\footnotesize(+12.15)}
& \cellcolor{bestgreen}\textbf{58.70} {\footnotesize(+2.40)}
& \cellcolor{bestgreen}\textbf{42.65} {\footnotesize(+6.35)}
& \cellcolor{whitebg}17.50 {\footnotesize(-0.26)}\\
\bottomrule
\end{tabular}
\caption{
Comparison of \textbf{Qwen2.5-7B-VL} and \textbf{Qwen2.5-3B-VL} under different training strategies, grouped by training dataset. Numbers are absolute accuracy (\%), with differences relative to the \emph{Vanilla} inference-only baseline shown in parentheses. \textbf{SFT} fine-tunes on HQ correct responses (using rejection-sampled model-generated reasonings), \textbf{VD-LF} constructs label-free HQ vs.\ LQ preferences, and \textbf{VD-LB} applies label-based VisualDeltas using HQ-correct vs.\ LQ-wrong pairs. Row and cell coloring: gray marks dataset groupings, blue marks VisualDelta-based results, dark green highlights best results, light green indicates positive gains, light red indicates negative changes, and dark red marks the largest drop.
}
\label{tab:main_results_table}
\end{table*}

\section{Experiments}
\label{sec:experiments}

\subsection{Experimental Settings}
\label{sec:exp_settings}

We evaluate VisualDeltas on five multimodal QA benchmarks: HiTab~\citep{cheng2022hitab}, WikiTQ~\citep{pasupat2015wikitable}, VQA v2~\citep{goyal2017vqa}, GQA~\citep{hudson2019gqa}, and MathVision~\citep{wang2024mathvision}. These datasets cover diverse capabilities, from hierarchical table reasoning to open-ended visual understanding. We use Qwen2.5-VL-7B and Qwen2.5-VL-3B as base models~\citep{qwen2_5_vl}.
For resolution perturbation, we downsample images to 10\% of original dimensions ($\alpha=0.1$) as the default LQ view.

\paragraph{Evaluated Methods.}
Figure~\ref{fig:method_overview} illustrates the VisualDeltas framework. We compare four strategies:
\begin{itemize}
    \item \textbf{Vanilla}: Inference only, no training.
    \item \textbf{SFT}: Rejection-sampling Fine-tuning on correct HQ responses only, where the training data consists of HQ image inputs and their corresponding rejection-sampled model-generated reasonings.
    \item \textbf{VD-LF (Label-free)}: VisualDeltas constructs preference pairs using \emph{all} HQ vs.\ LQ responses, regardless of their correctness. This variant requires no ground-truth labels.
    \item \textbf{VD-LB (Label-based)}: VisualDeltas with correctness filtering, which constructs preference pairs only from HQ-correct vs.\ LQ-wrong responses, requiring ground-truth labels for filtering.
\end{itemize}
All DPO training conditions exclusively on HQ inputs.

\subsection{Main Results: Effectiveness of VisualDeltas}
\label{sec:main_results}

Table~\ref{tab:main_results_table} summarizes the main results across datasets, model scales, and training strategies. Overall, VisualDeltas-based training leads to consistent improvements over inference-only baselines and SFT, while exhibiting more stable behavior under cross-dataset evaluation.

\paragraph{VisualDeltas vs.\ SFT.}
We first compare VisualDeltas with SFT, where both methods leverage HQ inputs during training. Across diverse combinations of training datasets and evaluation benchmarks, the label-based variant VD-LB generally yields stronger overall performance than SFT. While SFT often improves accuracy on the in-domain training dataset, it frequently exhibits noticeable performance degradation on out-of-domain benchmarks. For example, on the 7B model trained on VQA, SFT improves GQA by $+2.15\%$ but degrades MathVision by $-6.51\%$. In contrast, VD-LB achieves a larger gain on GQA ($+4.80\%$) while maintaining near-neutral performance on MathVision ($+0.49\%$). Similar trends are observed when training on GQA, suggesting that preference-based alignment better preserves transferable visual reasoning capabilities than direct supervision on correct outputs.

\paragraph{Effectiveness of label-free VisualDeltas.}
Even without correctness annotations, the label-free variant VD-LF achieves performance comparable to, and in many cases exceeding, supervised SFT. In particular, on WikiTQ with the 3B model, VD-LF yields substantial improvements when trained on either VQA or GQA, surpassing the corresponding gains obtained by SFT. More broadly, VD-LF demonstrates consistent advantages on table-centric and visually grounded reasoning tasks, indicating that relative visual quality signals induced by resolution differences provide an informative and reliable learning signal. These results suggest that effective preference alignment can be achieved without explicit correctness labels, enabling broader applicability in settings where annotation is unavailable or unreliable.

\paragraph{Impact of label-based filtering.}
Incorporating correctness information through label-based filtering (VD-LB) further strengthens performance relative to VD-LF, although the margin is typically modest. The benefits of VD-LB are most apparent on tasks where visual degradation directly affects reasoning outcomes, such as table understanding in WikiTQ. For example, on the 7B model trained on VQA, VD-LB improves WikiTQ accuracy while VD-LF exhibits a slight regression. Nevertheless, across most benchmarks, the difference between VD-LB and VD-LF remains small (generally within $0.3\%$--$1.5\%$), indicating that label-free preference signals already capture the majority of the performance gains, with label-based filtering providing incremental refinement.

\paragraph{Scalability across model sizes.}
VisualDeltas exhibits consistent effectiveness across model scales. Both 3B and 7B models benefit from preference-based training on all evaluated datasets, with performance gains remaining stable as model capacity increases. Improvements are observed across WikiTQ, VQA, and GQA, without introducing scale-dependent instability. Importantly, neither model scale displays the pronounced cross-dataset degradation commonly observed with SFT, suggesting that the proposed approach generalizes robustly across capacity regimes.

\paragraph{Summary.}
Taken together, these results demonstrate that VisualDeltas provides a robust alternative to SFT for multimodal reasoning models. Preference-based alignment consistently improves performance while substantially mitigating cross-dataset degradation. Moreover, label-free construction captures most of the benefits, enabling effective training without reliance on correctness annotations, and label-based filtering offers additional gains when such information is available.

\subsection{Generalization to Low-Quality Inputs}
\label{sec:lq_generalization}

A natural question arises: does VisualDeltas training on HQ images also improve performance when testing on degraded, LQ inputs? To evaluate this, we directly test all trained models on LQ versions of the test sets.

\begin{table}[!ht]
\centering
\scriptsize
\setlength{\tabcolsep}{3.5pt}
\renewcommand{\arraystretch}{0.95}
\resizebox{\columnwidth}{!}{
\begin{tabular}{lccccc}
\toprule
\multirow{2}{*}{Method} & \multicolumn{5}{c}{Test on LQ (0.1$\times$ Res)} \\
\cmidrule(lr){2-6}
& HiTab & WikiTQ & VQA & GQA & MathVision \\
\midrule
\multicolumn{6}{l}{\textbf{Qwen2.5-7B-VL}} \\
\midrule
\textit{No Training} & 33.46 & 40.00 & 50.55 & 33.40 & 23.95 \\
\midrule
\rowcolor{grouprow}
\textit{Train: HiTab} & & & & & \\
SFT & 34.34 & 45.35 & 51.20 & 32.30 & 18.55 \\
\rowcolor{vdrow}
VD-LF & 36.43 & 47.95 & 52.55 & \cellcolor{bestgreen}\textbf{36.80} & 22.37 \\
\rowcolor{vdrow}
VD-LB & \cellcolor{bestgreen}\textbf{37.44} & \cellcolor{bestgreen}\textbf{50.95} & \cellcolor{bestgreen}\textbf{55.05} & 36.65 & \cellcolor{bestgreen}\textbf{24.24} \\
\midrule
\rowcolor{grouprow}
\textit{Train: VQA} & & & & & \\
SFT & 29.42 & 43.30 & 55.35 & 34.35 & 16.84 \\
\rowcolor{vdrow}
VD-LF & \cellcolor{bestgreen}\textbf{34.22} & 45.70 & 55.35 & 38.70 & \cellcolor{bestgreen}\textbf{24.24} \\
\rowcolor{vdrow}
VD-LB & 34.09 & \cellcolor{bestgreen}\textbf{47.55} & \cellcolor{bestgreen}\textbf{59.10} & \cellcolor{bestgreen}\textbf{39.35} & 23.85 \\
\midrule
\rowcolor{grouprow}
\textit{Train: GQA} & & & & & \\
SFT & 31.88 & 45.40 & 54.40 & 36.20 & 15.89 \\
\rowcolor{vdrow}
VD-LF & 34.34 & \cellcolor{bestgreen}\textbf{48.20} & \cellcolor{bestgreen}\textbf{56.25} & 36.80 & 23.78 \\
\rowcolor{vdrow}
VD-LB & \cellcolor{bestgreen}\textbf{34.72} & \cellcolor{bestgreen}\textbf{48.20} & 55.60 & \cellcolor{bestgreen}\textbf{41.15} & \cellcolor{bestgreen}\textbf{24.18} \\
\bottomrule
\end{tabular}
}
\caption{
\textbf{Generalization to low-quality image inputs (Qwen2.5-7B-VL).} All models are trained on HQ images and tested on LQ images. \textbf{SFT} fine-tunes on HQ correct responses, \textbf{VD-LF} constructs label-free HQ vs.\ LQ preferences, and \textbf{VD-LB} applies label-based filtering using HQ-correct vs.\ LQ-wrong pairs. Best results per column are highlighted in bold with dark green background. Row and cell coloring: gray marks dataset groupings, blue marks VisualDeltas results, and dark green indicates best performance.
}
\label{tab:lq_generalization}
\end{table}

Table~\ref{tab:lq_generalization} evaluates model performance when trained on HQ images but tested on LQ images. While absolute accuracy drops across all tasks, the results reveal clear differences in how training strategies handle degraded visual inputs.

\paragraph{Amplified gains on structure-sensitive tasks under LQ evaluation.}
Comparing HQ and LQ results, VisualDeltas shows amplified improvements on layout- and structure-intensive datasets. For example, when trained on HiTab, VD-LB achieves +10.95 absolute gain on WikiTQ under LQ testing (from baseline 40.00 to 50.95), compared to only +3.50 gain under HQ testing. Similar amplification occurs on GQA, where VD-LF trained on HiTab achieves +3.40 gain under LQ versus +2.10 under HQ. In contrast, on less structure-dependent tasks like VQA, improvements remain more moderate (e.g., +8.55 for VD-LB trained on VQA, from baseline 50.55 to 59.10). This suggests VisualDeltas is particularly effective at reinforcing robust visual reasoning strategies that compensate for structural ambiguity.

\paragraph{Effectiveness of label-free and label-based VisualDeltas.}
Both VD-LF and VD-LB substantially outperform SFT under LQ evaluation, demonstrating that preference learning from visual quality variations is far more effective than correctness-only supervision for robustness. Between the two, VD-LB generally achieves stronger performance in most settings, while VD-LF remains competitive and occasionally achieves slightly better results on specific transfers (e.g., VQA-trained VD-LF on HiTab: 34.22\% vs. 34.09\%). This indicates that label-free preferences already capture most of the robustness gains, and label-based filtering provides additional refinement through cleaner supervision signals.

\paragraph{SFT fails to transfer under degradation.}
SFT exhibits severe instability when transferring from HQ training to LQ testing. In cross-dataset scenarios, SFT often performs worse than the no-training baseline. For instance, when trained on VQA, SFT achieves only 29.42\% on HiTab under LQ testing---below the baseline 33.46\%. Similar degradation occurs on MathVision, where VQA-trained SFT drops to 16.84\% versus the 23.95\% baseline. In contrast, VisualDeltas consistently outperforms both SFT and the baseline, with VD-LF trained on VQA achieving 34.22\% on HiTab and 24.24\% on MathVision. This demonstrates that correctness-only supervision overfits to high-fidelity visual features and fails to develop robustness to degraded inputs.

\paragraph{Summary.}
LQ evaluation reveals that VisualDeltas develops genuine robustness to degraded visual inputs, whereas SFT's correctness-only supervision creates brittle models that collapse when visual fidelity drops. Both label-free and label-based variants substantially outperform SFT, with VD-LB providing incremental improvements through cleaner supervision signals. This confirms that preference learning from visual quality variations is far more effective than correctness-only supervision for robustness.

\subsection{Qualitative Analysis: When Visual Information Matters}
\label{sec:qualitative_analysis}

To better understand \textit{what} VisualDeltas is improving, we analyze how model performance changes as input resolution decreases, and compare behaviors across datasets with different degrees of visual complexity. This analysis highlights a consistent pattern: VisualDeltas brings the most benefit when success depends on extracting and grounding \textbf{rich visual cues}, whereas tasks with relatively \textbf{sparse visual information} show little to no improvement.

\paragraph{VisualDeltas helps most on visually rich benchmarks.}
Across visually dense tasks---such as table understanding (HiTab, WikiTQ) and natural-image QA (VQA, GQA)---accuracy is strongly resolution-dependent: reducing resolution removes fine-grained structure (e.g., cell boundaries, text, object attributes/relations), and performance drops sharply (Fig.~\ref{fig:resolution_datasets}). In these settings, VisualDeltas can leverage the systematic gap between high-quality and low-quality inputs to provide informative preference signals that reinforce behaviors tied to accurate visual grounding.

\paragraph{MathVision behaves differently: little gain and near resolution invariance at inference.}
In contrast, MathVision is visually simpler (often dominated by clean expressions/diagrams; see Fig.~\ref{fig:dataset_examples} for examples) and relies comparatively more on the model's internal symbolic/multi-step reasoning than on extracting dense visual details. Empirically, the resolution--accuracy curve for MathVision is nearly flat, while other datasets show clear sensitivity to resolution (HiTab/WTQ strongly, VQA/GQA moderately; see Fig.~\ref{fig:resolution_datasets}). This inference-time invariance indicates that, for MathVision, reducing resolution does not substantially remove the information needed to solve the task, unlike perception-heavy benchmarks.

\paragraph{Implication: VisualDeltas improves visual information utilization, not general reasoning.}
Consistent with the resolution analysis, overall gains on MathVision are minimal: for example, on the same 7B target model, VisualDeltas changes MathVision accuracy only marginally relative to baseline (e.g., +0.26 to +0.29 in the reported setting), whereas other datasets show larger improvements. Taken together, these observations suggest that VisualDeltas primarily strengthens the model's ability to \textit{use visual evidence effectively}---i.e., robust visual grounding and perception-conditioned reasoning---rather than improving reasoning capability in a task-agnostic sense.

\begin{figure}[ht!]
    \centering
    \includegraphics[width=0.38\textwidth]{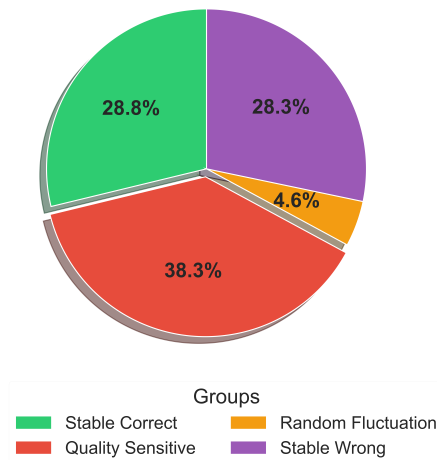}
    \caption{\textbf{Sample category distribution on HiTab.} Quality-Sensitive samples (HQ correct, LQ wrong) comprise 38.3\% of the dataset, providing clean HQ$\succ$LQ preference pairs for VD-LB training.}
    \label{fig:stability_distribution}
\end{figure}

\subsection{What Makes VisualDeltas Effective?}
\label{sec:what_makes_vd_effective}

We analyze two factors that contribute to VisualDeltas' effectiveness: sample category distribution under resolution perturbation, and behavioral differences between HQ and LQ responses.

\subsubsection{Label-Based Filtering Selects High-Quality Pairs}

Why does VD-LB achieve strong performance with relatively few training samples? We analyze HiTab samples by correctness under HQ and LQ views. Always-Correct samples are correct at both resolutions. Quality-Sensitive samples are correct at HQ but wrong at LQ. Always-Wrong samples fail at both. Paradoxically-Robust samples are wrong at HQ but correct at LQ. This fourth category is rare and likely reflects estimation errors.

Our analysis reveals that 38.3\% of HiTab samples (2,841 out of 7,417) are Quality-Sensitive (Fig.~\ref{fig:stability_distribution}). These naturally form high-quality preference pairs where HQ is strictly better than LQ. By leveraging labels to select exactly this category, VD-LB concentrates training on the most informative samples. This label-driven filtering makes preference learning more efficient than using all pairs indiscriminately.

Compared to label-free VD-LF, which must learn from noisy mixed-quality pairs, VD-LB benefits from precise selection. Quality-Sensitive samples directly capture the causal mechanism: visual degradation causes reasoning failure. Always-Correct samples offer weaker signals since both responses are acceptable. Always-Wrong samples provide no useful information. By filtering to Quality-Sensitive pairs only, VD-LB achieves strong results with approximately 2,800 training samples. Correctness labels enable more efficient preference learning by identifying the most informative comparisons.

\subsubsection{LQ Responses Exhibit Compensatory Inefficiency}

\begin{figure}[ht!]
    \centering
    \includegraphics[width=0.45\textwidth]{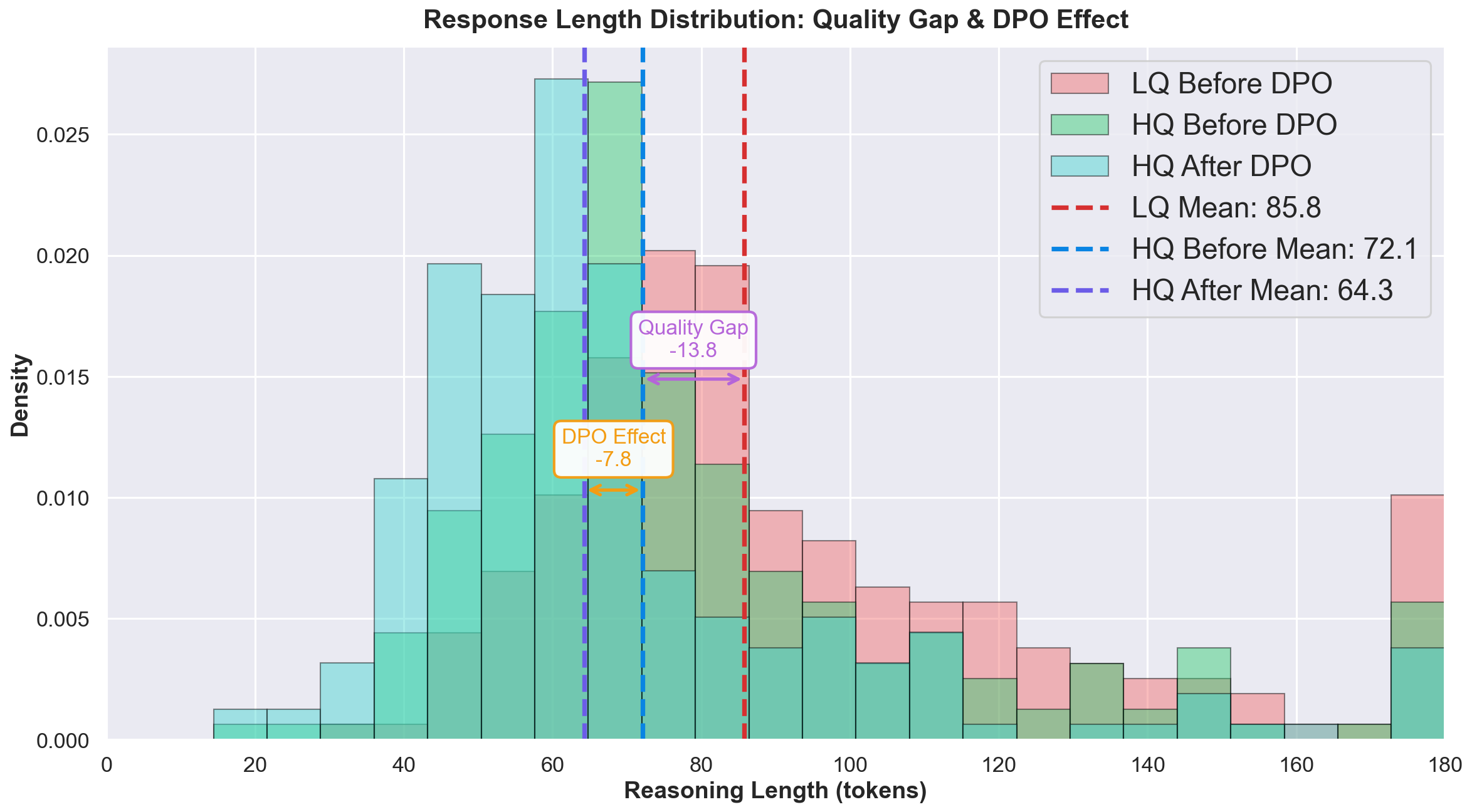}
    \caption{\textbf{Response length distribution by category.} LQ responses are consistently longer than HQ responses, suggesting that degraded visual inputs trigger compensatory but ineffective reasoning. After DPO training on VisualDeltas preference pairs, the HQ distribution shifts left and becomes sharper, with reduced mean and median token lengths. This demonstrates that DPO improves reasoning efficiency---models learn to produce more concise responses while maintaining higher accuracy.}
    \label{fig:response_length}
\end{figure}

Resolution degradation also affects reasoning behavior. Figure~\ref{fig:response_length} compares response lengths across categories. A striking pattern emerges: LQ responses are consistently longer than HQ responses, particularly for Quality-Sensitive samples.

This reveals a key insight: when visual information is unclear, the model attempts to compensate by generating longer, more verbose responses. However, this verbosity is ineffective---the model produces more tokens but achieves lower accuracy. For Quality-Sensitive samples, LQ elicits longer incorrect responses while HQ yields concise correct answers. Degraded inputs trigger \emph{compensatory inefficiency}: the model works harder but achieves worse outcomes.

This behavioral difference provides a strong preference learning signal. HQ responses are not only more accurate but also more efficient. DPO leverages this signal to shift the token distribution---the distribution becomes sharper and left-skewed, with reduced mean and median token lengths. Crucially, this increased conciseness coincides with higher accuracy, demonstrating improved reasoning efficiency. DPO suppresses verbose, ineffective patterns triggered by degraded inputs and reinforces compact, efficient reasoning grounded in clear visual perception.

\subsection{VisualDeltas Generalizes to Diverse Perturbation Types}
\label{sec:perturbation_comparison}

While our primary experiments use \textbf{Resolution Reduction} as the degradation mechanism, VisualDeltas is a general framework that can leverage various types of visual quality perturbations. We validate this by comparing resolution reduction against two alternative degradation strategies: \textbf{Gaussian Noise} (adding random pixel noise) and \textbf{Motion Blur} (simulating motion artifacts).

\paragraph{All perturbation types yield comparable gains.}
Across five multimodal QA benchmarks, we observe that all three perturbation strategies achieve similar improvements over the baseline. Resolution reduction (our default VD-LF and VD-LB methods) performs competitively on most datasets, while Gaussian Noise and Motion Blur provide comparable gains. This consistency demonstrates that VisualDeltas is not tied to a specific degradation type---rather, the key principle is creating a meaningful quality gap between HQ and LQ views that systematically exposes reasoning brittleness.

\paragraph{Practical advantages of resolution reduction.}
Among the three strategies, resolution reduction offers several practical benefits: (1) it is computationally cheap, requiring only a single resize operation; (2) it is deterministic and reproducible across runs; (3) it works universally across all image types without parameter tuning; and (4) it naturally occurs in real-world scenarios (e.g., low-resolution images, compression artifacts). These properties make resolution reduction particularly attractive as the default perturbation mechanism for VisualDeltas.

\paragraph{Implications for framework generality.}
The consistent performance across diverse perturbation types suggests that VisualDeltas can exploit a wide range of quality-induced variations. This opens the door to extending the framework to other degradation types (e.g., JPEG compression, occlusion, color distortion) and to combining multiple perturbations for stronger supervision signals. Detailed experimental results and comparisons are provided in Appendix~\ref{appendix:perturbation_comparison}.

\section{Conclusion}
\label{sec:conclusion}

VisualDeltas leverages resolution-induced reasoning deltas as preference signals. By contrasting responses under high-quality and low-quality views, it generates supervision without external annotations. Experiments show consistent improvements over SFT in accuracy and generalization.

\section*{Impact Statement}

This work introduces VisualDeltas, a data-efficient preference learning framework for multimodal question answering that relies on intrinsically constructed supervision induced by visual quality perturbations.

\paragraph{Societal Impact.}
By reducing dependence on large-scale human annotations or external preference models, VisualDeltas lowers the cost of training robust multimodal systems and may benefit low-resource research settings and real-world applications with imperfect visual inputs, such as document understanding under low resolution.

\paragraph{Ethical Considerations.}
The preference signals used in this work are heuristic and derived from model behavior under resolution changes, which may be noisy or unreliable in some cases.
As a result, the method may amplify existing biases in the pretrained model if applied without care.
VisualDeltas is intended as a lightweight training mechanism and should not replace human oversight in high-stakes applications.


\bibliography{example_paper}
\bibliographystyle{icml2026}

\appendix
\onecolumn



\section{Datasets and Evaluation Benchmarks}
\label{appendix:datasets_info}

We consider a diverse collection of multimodal question answering benchmarks spanning structured tables and natural images.
Specifically, we include HiTab, WikiTableQuestions (WikiTQ), VQA v2, GQA, and MathVision.
These datasets differ in visual modality, question formulation, and reasoning complexity, providing complementary coverage of multimodal reasoning abilities.

\paragraph{HiTab}
HiTab is a large-scale benchmark for hierarchical table question answering introduced by Microsoft Research.
Each example consists of a table rendered as an image and a natural language question that requires multi-step reasoning over hierarchical table structures, such as aggregation, comparison, sorting, and navigation across grouped rows and columns.
Compared to flat table QA datasets, HiTab emphasizes compositional reasoning over complex table layouts.
We follow the standard exact-match accuracy metric for evaluation.

\paragraph{WikiTableQuestions}
WikiTableQuestions (WikiTQ) is a benchmark for question answering over tables extracted from Wikipedia.
Questions typically involve factual lookup, comparison, and lightweight aggregation over tabular entries.
Although originally proposed as a text-based dataset, WikiTQ has been widely adopted in multimodal settings by rendering tables as images.
We use the standard dataset splits and evaluation protocol provided by the benchmark.

\paragraph{VQA}
The Visual Question Answering (VQA v2) dataset consists of natural images paired with open-ended questions that require visual understanding and reasoning.
Questions cover a broad range of skills, including object recognition, attribute identification, spatial reasoning, and counting.
Evaluation is performed using the official VQA accuracy metric.

\paragraph{GQA}
GQA is a compositional visual reasoning benchmark designed to evaluate multi-step reasoning and systematic generalization.
Questions are generated from scene graphs and often require explicit reasoning chains involving object attributes, relations, and logical operations.
We follow the standard accuracy metric provided by the dataset.

\paragraph{MathVision}
MathVision is a multimodal benchmark focusing on mathematical reasoning over visually presented content, including tables, charts, diagrams, and mathematical expressions.
Questions typically require symbolic manipulation, numerical reasoning, or multi-step logical inference grounded in visual inputs.
We follow the official evaluation protocol of the benchmark.

\paragraph{Data Preprocessing and Sampling}
For HiTab and WikiTableQuestions, which are originally provided as structured tabular data (JSON/CSV format), we convert tables to images using custom rendering scripts to enable multimodal processing.
For all datasets except HiTab, we subsample 8,000 training examples and 2,000 test examples for experimental evaluation.
For HiTab, we use the full dataset consisting of 7,417 training examples and 1,584 test examples.

\begin{table}[ht]
\centering
\small
\begin{tabular}{lcccc}
\toprule
Dataset & Visual Modality & Task Type & Train Size & Test Size \\
\midrule
HiTab &
Table images &
Hierarchical table QA &
7,417 &
1,584 \\
WikiTableQuestions &
Table images &
Table QA &
8,000 &
2,000 \\
VQA v2 &
Natural images &
Open-ended VQA &
8,000 &
2,000 \\
GQA &
Natural images &
Compositional VQA &
8,000 &
2,000 \\
MathVision &
Mixed visual formats &
Visual math reasoning &
8,000 &
2,000 \\
\bottomrule
\end{tabular}
\caption{
Summary of datasets used in this work.
Training is conducted on HiTab, VQA, and GQA, while evaluation is performed on all datasets.
For HiTab and WikiTableQuestions, tables are converted from JSON/CSV format to images.
}
\label{tab:dataset_summary}
\end{table}

\paragraph{Dataset Usage and Rationale}
Among the five datasets, we select HiTab, VQA, and GQA for training, and evaluate on all datasets.
This choice is motivated by the complementary reasoning capabilities represented by these benchmarks.
HiTab focuses on visually grounded table question answering and captures structured, layout-sensitive reasoning.
VQA represents general-purpose visual understanding over natural images, while GQA emphasizes compositional and relational reasoning with explicit multi-step dependencies.
Together, these datasets span the core multimodal abilities that we aim to model: table-based reasoning, general visual perception, and compositional reasoning.
Evaluation on WikiTQ and MathVision further assesses generalization to unseen table structures and visually grounded mathematical reasoning.

\begin{figure*}[!t]
    \centering
    \begin{tabular}{cc}
        \includegraphics[width=0.32\linewidth]{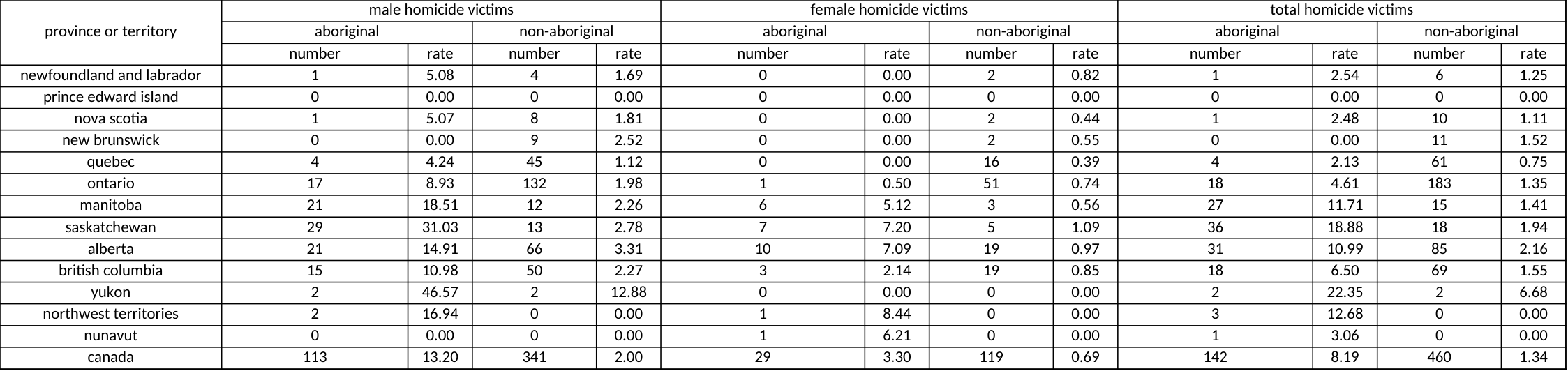} &
        \includegraphics[width=0.32\linewidth]{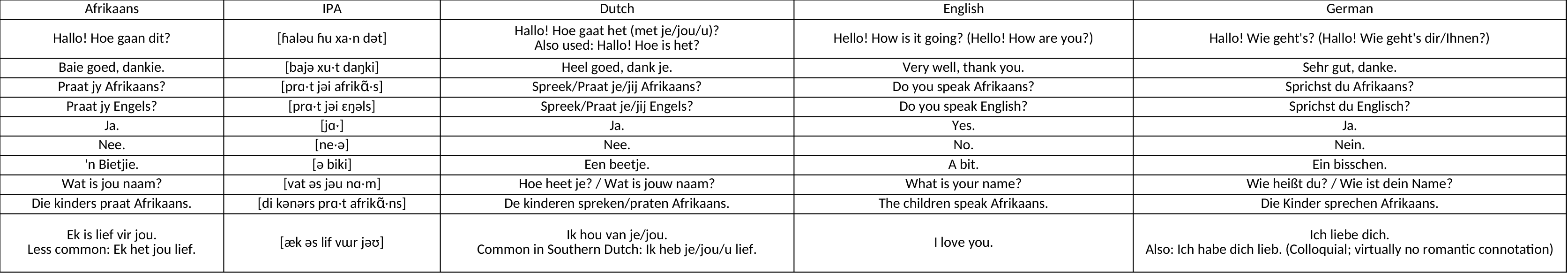} \\
        \small{(a) HiTab: Hierarchical table with complex layout} &
        \small{(b) WikiTableQuestions: Structured table} \\[0.3cm]
        \includegraphics[width=0.32\linewidth]{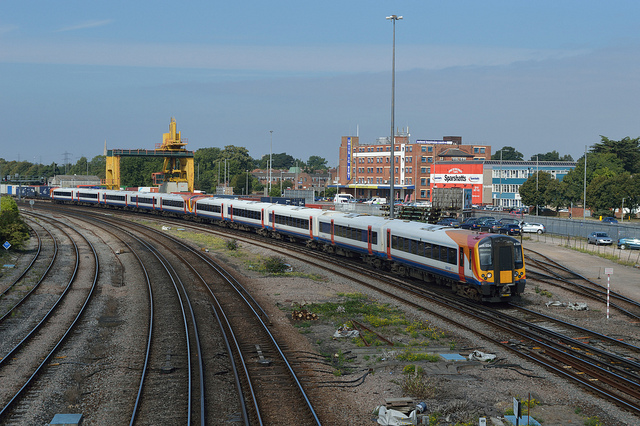} &
        \includegraphics[width=0.32\linewidth]{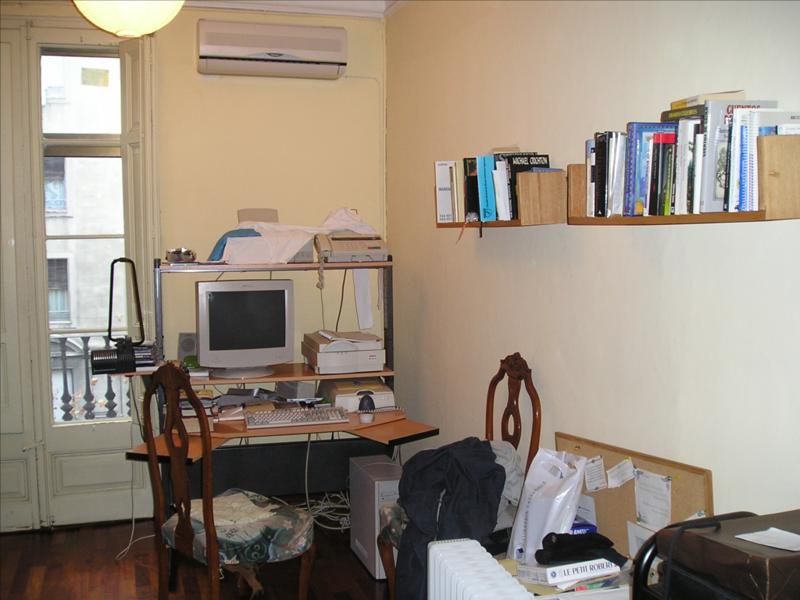} \\
        \small{(c) VQA: Natural image with question} &
        \small{(d) GQA: Compositional scene reasoning} \\[0.3cm]
        \multicolumn{2}{c}{\includegraphics[width=0.32\linewidth]{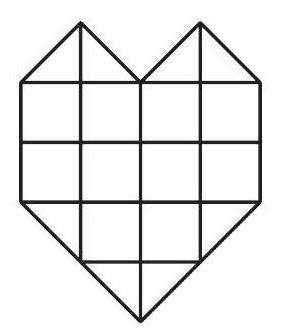}} \\
        \multicolumn{2}{c}{\small{(e) MathVision: Simple mathematical expressions}} \\
    \end{tabular}
    \caption{
    \textbf{Visual Examples from Different Datasets.} (a)-(b) Table datasets (HiTab, WikiTableQuestions) feature complex, dense structures requiring precise visual grounding. (c)-(d) Natural image datasets (VQA, GQA) contain real-world scenes with varying complexity. (e) MathVision contains simple mathematical expressions and diagrams that rely more on reasoning than visual details. This visual complexity hierarchy explains why HiTab/WTQ show strong resolution dependence while MathVision remains resolution-independent.
    }
    \label{fig:dataset_examples}
\end{figure*}

\paragraph{Visual Examples}
Figure~\ref{fig:dataset_examples} shows representative visual examples from each dataset, illustrating the varying levels of visual complexity that explain the differential resolution sensitivity observed.

\begin{figure*}[!t]
    \centering
    \includegraphics[width=0.95\linewidth]{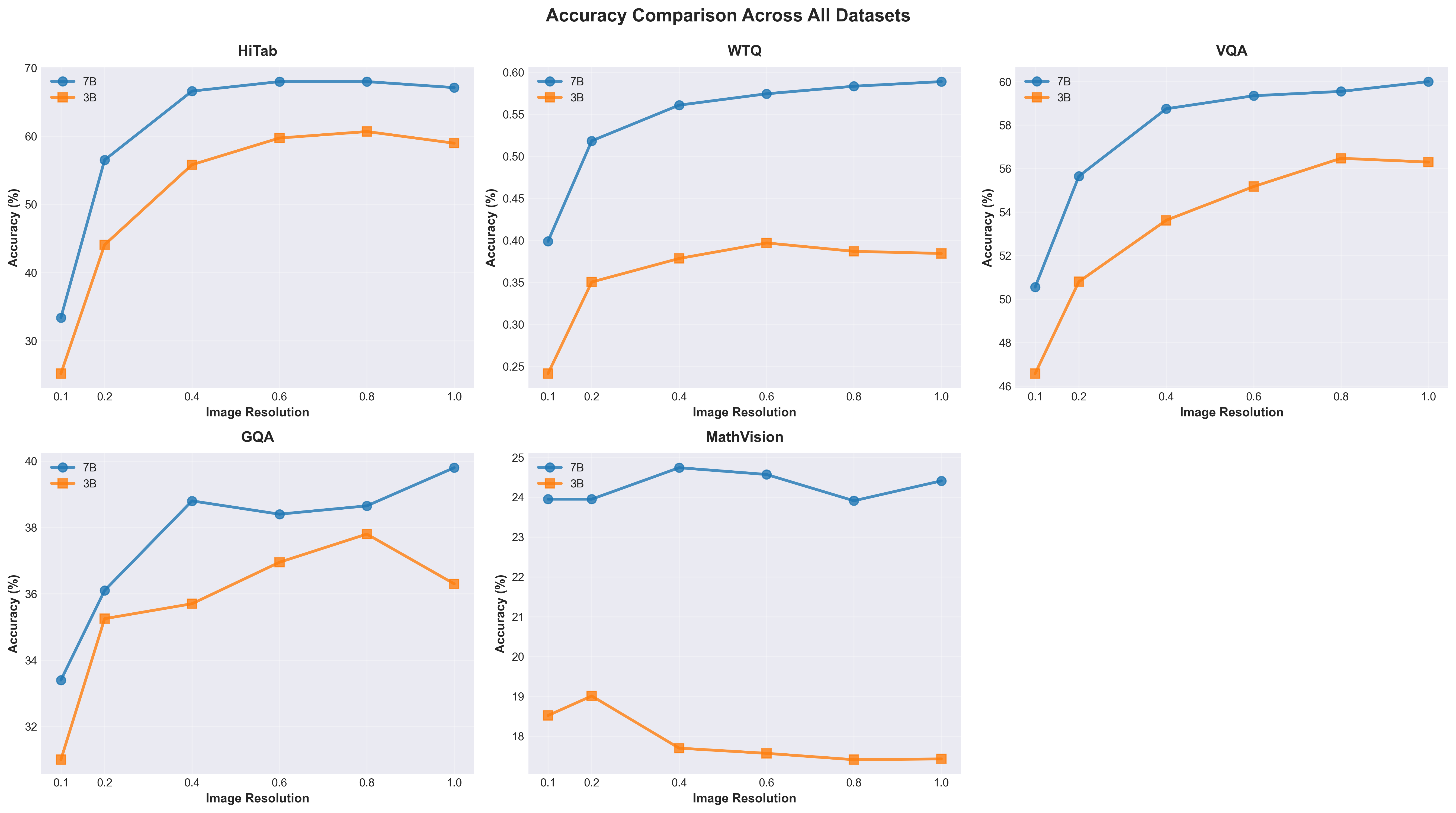}
    \caption{
    \textbf{Accuracy vs. Resolution Across Datasets.} Performance trends across different datasets under varying image resolutions. HiTab and WTQ show strong resolution dependence (accuracy doubling from 10\% to 100\% resolution), reflecting their visual complexity with dense, structurally intricate tables. MathVision maintains nearly flat performance across all resolutions, indicating reliance on reasoning rather than visual details. VQA and GQA show intermediate resolution sensitivity. This demonstrates that VisualDeltas specifically targets visually grounded reasoning tasks.
    }
    \label{fig:resolution_datasets}
\end{figure*}

\paragraph{Resolution Analysis Across Datasets}
Figure~\ref{fig:resolution_datasets} shows performance trends across different datasets under varying image resolutions.

\section{Impact of Image Resolution on Model Inference and Reasoning}
\label{appendix:compression_and_reasoning}

This appendix presents a comprehensive analysis of how image resolution reduction affects both the inference behavior and reasoning quality of the Qwen2.5-VL-7B model on the HiTab table understanding benchmark. We systematically evaluate model performance across six resolution levels (0\%, 20\%, 40\%, 60\%, 80\%, and 90\% dimension reduction) to characterize the trade-offs between image quality and model accuracy, while examining the internal reasoning dynamics that underlie these performance changes.

\subsection{Experimental Setup}
\label{subsec:exp_setup_merged}

\paragraph{Dataset and Model}
We conducted experiments on the HiTab dataset \citep{cheng2022hitab}, which contains 1,584 table-based question-answering samples spanning diverse domains including financial reports, scientific data, and statistical summaries. Each sample was processed at six different resolution levels, resulting in a total of 9,504 inference results. The model was configured to generate responses in a chain-of-thought format with explicit \texttt{<thinking>} and \texttt{<answer>} tags.

\paragraph{Resolution Perturbation Methodology}
Image resolution reduction was applied using PIL's resize function with different scaling factors. The compression rates (0\%, 20\%, 40\%, 60\%, 80\%, 90\%) represent the percentage by which the image dimensions were reduced. For example, 90\% compression means reducing the image to 10\% of its original dimensions (height and width each scaled to 10\%). This is achieved through PIL's bilinear interpolation, which simulates the effect of lossy compression through resolution degradation rather than JPEG artifact injection.

\paragraph{Evaluation Metrics}
\textbf{Exact Match (EM)} requires the predicted answer to exactly match the ground truth string after normalization. \textbf{Tolerance Match (TM)} allows $\pm 0.5$ tolerance for numerical answers to account for minor rounding differences arising from OCR-like digit recognition errors.

\subsection{Accuracy Analysis and Perceptual Thresholds}
\label{subsec:accuracy_thresholds}

\begin{figure}[htbp]
    \centering
    \includegraphics[width=0.7\textwidth]{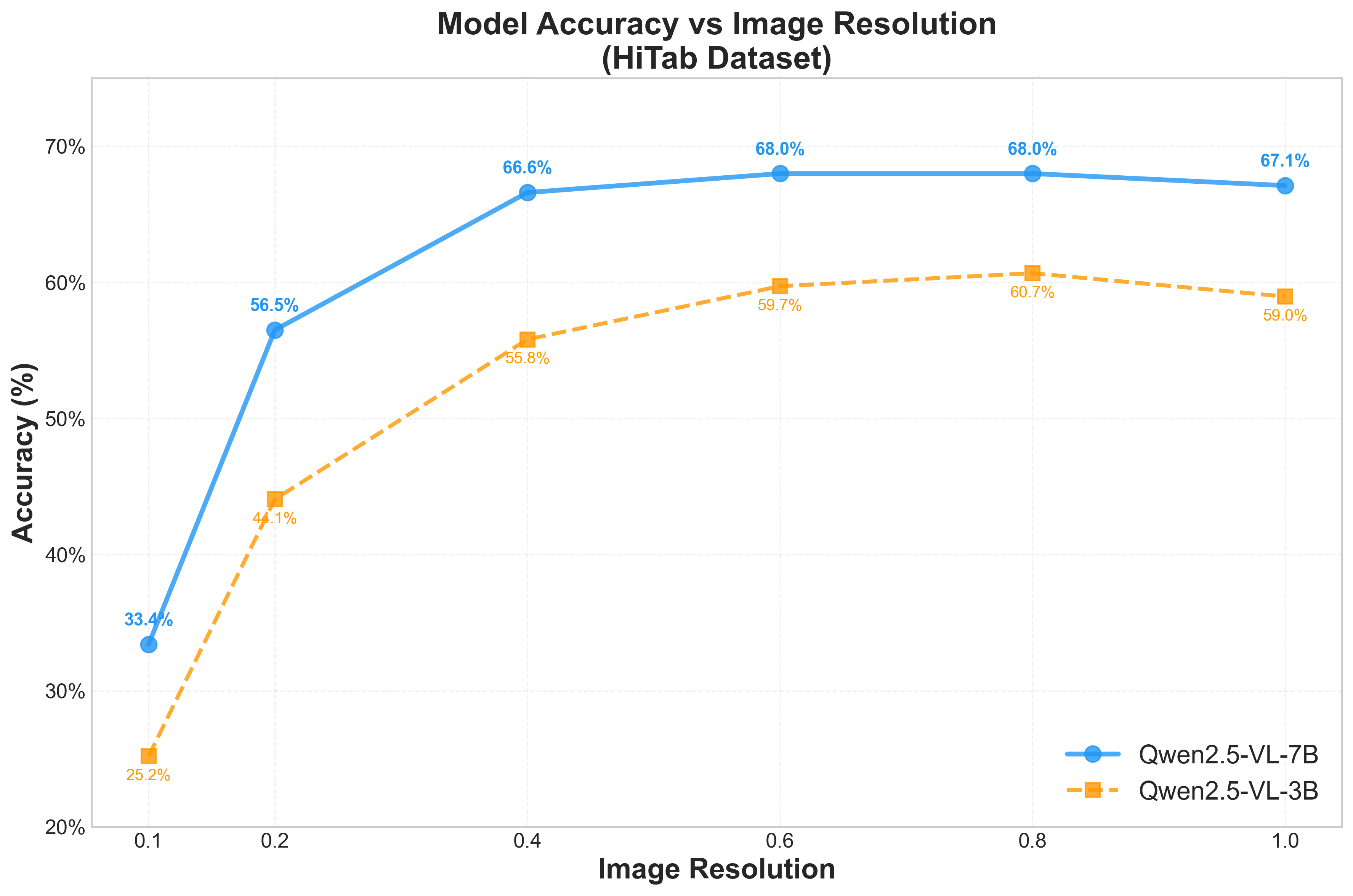}
    \caption{\textbf{Resolution-Accuracy Relationship.} The 7B model exhibits a characteristic three-phase pattern: stability at high resolution (100\%--40\%), rapid degradation in the critical zone (40\%--20\%), and severe failure at extreme low resolution (20\%--10\%). The 3B model shows a similar pattern with systematically lower accuracy, confirming the robustness of this phenomenon across model sizes.}
    \label{fig:accuracy_curve}
\end{figure}

Figure~\ref{fig:accuracy_curve} and Table~\ref{tab:accuracy_metrics} present the accuracy-compression relationship.

\begin{table}[htbp]
    \centering
    \caption{Accuracy metrics across resolution levels for Qwen2.5-VL-7B.}
    \label{tab:accuracy_metrics}
    \begin{tabular}{ccccc}
        \toprule
        \textbf{Resolution} & \textbf{Accuracy} & \textbf{Correct} & \textbf{Total} & \textbf{Change} \\
        \midrule
        100\% (1.0) & 67.11\% & 1063 & 1584 & --      \\
        80\%  (0.8) & 67.99\% & 1077 & 1584 & +0.88\% \\
        60\%  (0.6) & 67.99\% & 1077 & 1584 & +0.88\% \\
        40\%  (0.4) & 66.60\% & 1055 & 1584 & -0.51\% \\
        20\%  (0.2) & 56.50\% & 895  & 1584 & -10.61\% \\
        10\%  (0.1) & 33.40\% & 529  & 1584 & -33.71\% \\
        \bottomrule
    \end{tabular}
\end{table}

The accuracy curve reveals a characteristic three-phase degradation pattern:
\begin{itemize}
    \item \textbf{Stability phase} (100\%--40\% resolution): Accuracy remains stable (66.60\%--67.99\%), indicating the model's visual encoder possesses inherent robustness to moderate resolution reduction.
    \item \textbf{Critical transition phase} (40\%--20\% resolution): Accuracy drops from 66.60\% to 56.50\% (15.2\% relative degradation), suggesting a perceptual threshold beyond which resolution loss corrupts fundamental visual features.
    \item \textbf{Severe degradation phase} (20\%--10\% resolution): Accuracy plummets to 33.40\% (40.9\% relative decrease), with performance approaching random chance.
\end{itemize}

\paragraph{Model Size Comparison}
To assess scalability, we compare the 7B model against a 3B variant across the same resolution levels. The 3B model exhibits a similar degradation pattern but with systematically lower absolute accuracy: 58.96\% at 100\% resolution, 60.67\% at 80\%, 59.72\% at 60\%, 55.81\% at 40\%, 44.07\% at 20\%, and 25.19\% at 10\% resolution. The 7B model consistently outperforms the 3B model by 8--12 percentage points across all resolution levels, with the gap relatively stable (8.15\% at 100\%, 8.21\% at 10\%). This suggests that model scale improves absolute performance but does not significantly alter resolution sensitivity patterns. Both models show the characteristic three-phase degradation, confirming that this phenomenon is robust across model sizes.

\subsection{Sample Stability Classification}
\label{subsec:stability_classification}

We classify each sample based on its correctness pattern across high-resolution (0\% compression) and low-resolution (90\% compression) inputs, yielding four distinct quality groups.

\begin{figure}[htbp]
    \centering
    \includegraphics[width=0.48\textwidth]{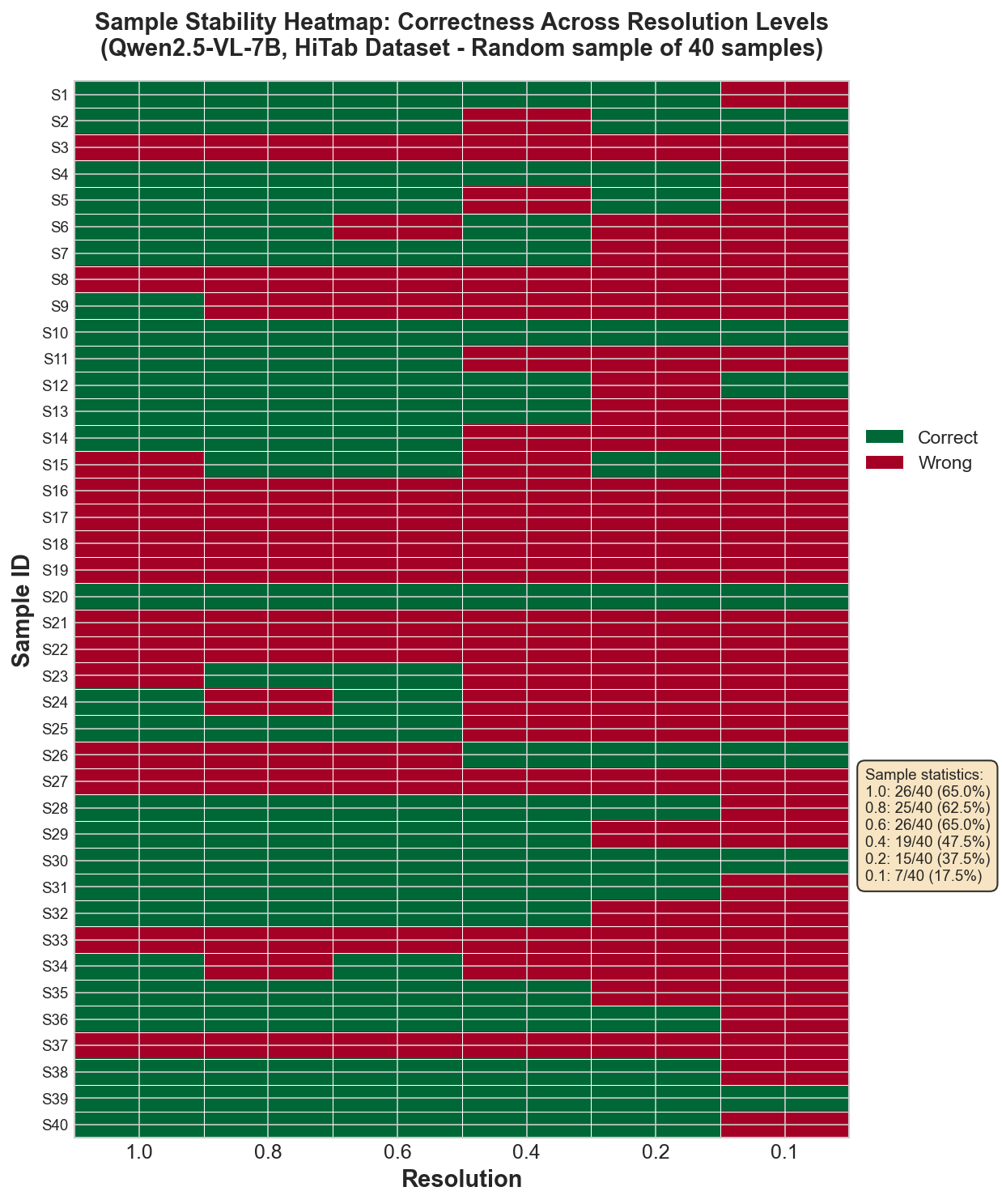}
    \caption{\textbf{Sample Stability Heatmap.} Failure transitions are unidirectional---once a question begins failing, it rarely recovers.}
    \label{fig:stability_heatmap}
\end{figure}

\begin{itemize}
    \item \textbf{Stable Correct (28.8\%)}: Correct at both high and low resolution levels (456 samples). These samples represent questions the model can answer reliably regardless of image quality.
    \item \textbf{Quality Sensitive (38.3\%)}: Correct at high resolution but wrong at low resolution (607 samples). These samples provide natural preference pairs for DPO training, where high-resolution outputs serve as preferred examples.
    \item \textbf{Random Fluctuation (4.6\%)}: Wrong at high resolution but correct at low resolution (73 samples). These represent fortunate guesses or noise in the low-resolution setting.
    \item \textbf{Stable Wrong (28.3\%)}: Fail at both high and low resolution levels (448 samples). These represent fundamental model limitations or ambiguous questions.
\end{itemize}

\section{Ablation Studies \& Robustness}
\label{appendix:hq_correct_vs_hq_wrong}

\subsection{Does the gain come from Correctness or Visual Delta?}
We compare VD-LB against a baseline DPO trained on ``HQ-Correct $\succ$ HQ-Wrong'' pairs (Table~\ref{tab:vds_vs_hq_main}). While HQ-only preference learning provides gains, VD-LB consistently outperforms it (e.g., \textbf{+10.23} cumulative gain vs. +8.34 on 7B). This confirms that the \textit{visual delta}---the contrast between robust perception and degraded perception---adds unique supervisory value beyond mere correctness.

\begin{table}[t]
\centering
\footnotesize
\setlength{\tabcolsep}{3pt}
\begin{tabular}{lcccc}
\toprule
Train Set & \shortstack{VD-S\\Total\\(7B)} & \shortstack{HQ vs. HQ\\Total\\(7B)} & \shortstack{VD-S\\Total\\(3B)} & \shortstack{HQ vs. HQ\\Total\\(3B)} \\
\midrule
HiTab & \textbf{+10.23} & +10.21 & \textbf{+10.10} & +9.42 \\
VQA   & \textbf{+10.39} & +8.34  & \textbf{+9.85}  & +8.50 \\
GQA   & \textbf{+12.84} & +9.12  & \textbf{+10.95} & +8.91 \\
\bottomrule
\end{tabular}
\caption{\textbf{Cumulative Accuracy Gains:} VD-LB vs. HQ-only DPO across all evaluation benchmarks. VD-LB achieves higher total gains than correctness-based DPO alone, demonstrating that visual quality deltas provide complementary supervision beyond mere correctness.}
\label{tab:vds_vs_hq_main}
\end{table}

\subsection{Sensitivity to Perturbation Strength}
We study the resolution downsampling ratio $\alpha$ (detailed in Appendix~\ref{appendix:resolution_perturbation_strength}). We find an inverted-U trend: gains are most pronounced at moderate to strong degradation ($\alpha=0.1$ or $0.2$). Overly mild perturbations ($\alpha \ge 0.6$) fail to induce sufficient behavioral divergence, confirming that a tangible ``quality gap'' is necessary for effective preference learning.

\subsection{Generalization to Other Degradations}
VisualDeltas is not limited to resolution. As discussed in Section~\ref{sec:perturbation_comparison}, applying \textbf{Gaussian Noise} or \textbf{Motion Blur} as the degradation operator yields comparable gains. This suggests our framework is a general recipe for exploiting perceptual robustness. Notably, VD-LB achieves competitive or better performance than noise-based methods on most benchmarks while avoiding the computational overhead of noise generation during inference.

\subsection{DPO with High-Quality Correct vs.\ High-Quality Wrong}

In this appendix, we present an ablation study isolating the contribution of correctness-based filtering from visual quality-induced deltas. Preference pairs are constructed from high-quality (HQ) visual inputs only, with the preference direction determined solely by answer correctness. This setting, \textbf{DPO (HQ Correct vs.\ HQ Wrong)}, allows us to disentangle the effects of visual quality differences and correctness signals.

Our main method, VD-S (HQ Correct vs.\ LQ Wrong), leverages two sources of supervision: a correctness signal and a visual quality delta. This naturally raises the question: \emph{how much of the observed performance gain can be attributed to correctness alone?} The HQ-only setting answers this.

\textbf{Experimental setup.} Multiple responses are sampled from the model for each HQ visual input. Preference pairs are constructed by selecting a correct response as preferred and an incorrect response as dispreferred. All other hyperparameters remain identical to the main experiments. This ablation is conducted on both Qwen2.5-7B-VL and Qwen2.5-3B-VL to assess scalability.

\subsection*{Comparison: VD-S vs HQ-only}

\begin{table*}[ht!]
\centering
\small
\setlength{\tabcolsep}{4pt}
\begin{tabular}{llcccccccc}
\toprule
Train & Method & \multicolumn{4}{c}{Qwen2.5-7B-VL} & \multicolumn{4}{c}{Qwen2.5-3B-VL} \\
\cmidrule(lr){3-6} \cmidrule(lr){7-10}
& & HiTab & WikiTQ & VQA & GQA & HiTab & WikiTQ & VQA & GQA \\
\midrule
\multirow{2}*{HiTab}
& VD-S       & \textbf{71.40} {\scriptsize(+3.47)} & \textbf{69.90} {\scriptsize(+3.50)} & \textbf{63.55} {\scriptsize(+3.55)} & \textbf{42.90} {\scriptsize(+2.95)} & \textbf{61.11} {\scriptsize(+2.15)} & 51.35 {\scriptsize(+12.90)} & \textbf{60.85} {\scriptsize(+4.55)} & 39.75 {\scriptsize(+3.45)} \\
& HQ vs.\ HQ & 71.15 {\scriptsize(+3.22)} & 68.95 {\scriptsize(+2.55)} & 63.15 {\scriptsize(+3.15)} & 40.90 {\scriptsize(+0.95)} & 60.67 {\scriptsize(+1.71)} & 50.05 {\scriptsize(+11.60)} & 61.50 {\scriptsize(+5.20)} & 39.80 {\scriptsize(+3.50)} \\
\midrule
\multirow{2}*{VQA}
& VD-S       & 70.58 {\scriptsize(+2.65)} & 68.15 {\scriptsize(+1.75)} & \textbf{68.20} {\scriptsize(+8.20)} & \textbf{44.75} {\scriptsize(+4.80)} & 59.09 {\scriptsize(+0.13)} & \textbf{52.45} {\scriptsize(+14.00)} & \textbf{61.45} {\scriptsize(+5.15)} & \textbf{42.50} {\scriptsize(+6.20)} \\
& HQ vs.\ HQ & \textbf{70.20} {\scriptsize(+2.27)} & \textbf{68.60} {\scriptsize(+2.20)} & 66.55 {\scriptsize(+6.55)} & 43.65 {\scriptsize(+3.70)} & 61.24 {\scriptsize(+2.28)} & 50.60 {\scriptsize(+12.15)} & 61.90 {\scriptsize(+5.60)} & 41.35 {\scriptsize(+5.05)} \\
\midrule
\multirow{2}*{GQA}
& VD-S       & 69.19 {\scriptsize(+1.26)} & 69.60 {\scriptsize(+3.20)} & \textbf{66.00} {\scriptsize(+6.00)} & \textbf{47.50} {\scriptsize(+7.55)} & \textbf{61.36} {\scriptsize(+2.40)} & 50.60 {\scriptsize(+12.15)} & \textbf{58.70} {\scriptsize(+2.40)} & \textbf{42.65} {\scriptsize(+6.35)} \\
& HQ vs.\ HQ & \textbf{70.45} {\scriptsize(+2.52)} & \textbf{70.35} {\scriptsize(+3.95)} & 64.35 {\scriptsize(+4.35)} & 45.25 {\scriptsize(+5.30)} & 59.53 {\scriptsize(+0.57)} & \textbf{48.90} {\scriptsize(+10.40)} & 59.65 {\scriptsize(+3.35)} & 43.15 {\scriptsize(+6.85)} \\
\bottomrule
\end{tabular}
\caption{Comparison between \textbf{VD-S (HQ Correct vs.\ LQ Wrong)} and \textbf{DPO (HQ Correct vs.\ HQ Wrong)} across Qwen2.5-7B-VL and 3B-VL. Bold indicates the best score per model for each metric. Numbers in parentheses indicate absolute improvements relative to the respective baseline.}
\label{tab:vd_s_vs_hq_merged}
\end{table*}

\textbf{Analysis.} Across both scales, VD-S generally outperforms HQ-only, with particularly clear advantages on in-domain datasets (HiTab, VQA, GQA). The visual quality delta provides complementary supervision, exposing structured errors that correctness alone cannot.

\subsection*{Aggregate Accuracy Improvements}

\begin{table*}[ht!]
\centering
\small
\setlength{\tabcolsep}{6pt}
\begin{tabular}{lcccc}
\toprule
Train Set & VD-S Total Diff (7B) & HQ vs.\ HQ Total Diff (7B) & VD-S Total Diff (3B) & HQ vs.\ HQ Total Diff (3B) \\
\midrule
HiTab & \textbf{+10.23} & +10.21 & \textbf{+10.10} & +9.42 \\
VQA   & \textbf{+10.39} & +8.34  & \textbf{+9.85}  & +8.50 \\
GQA   & \textbf{+12.84} & +9.12  & \textbf{+10.95} & +8.91 \\
\bottomrule
\end{tabular}
\caption{Summed absolute accuracy improvements across all evaluation benchmarks, for both 7B and 3B models, relative to the baseline.}
\label{tab:total_gain_merged}
\end{table*}

\textbf{Analysis.} Aggregated trends reinforce the complementary nature of correctness and visual quality signals. Correctness-based DPO alone consistently improves performance across both model scales, while VD-S provides additional gains, particularly on visually complex datasets such as GQA. Cross-scale trends are highly consistent, demonstrating scalability and robustness of the preference learning approach.

\textbf{Summary.} Across both 7B and 3B models:

\begin{itemize}
    \item Correctness alone is a strong supervision signal, effective even for smaller models.  
    \item VD-S generally achieves the largest improvements, highlighting the importance of visual quality deltas.  
    \item Trends are consistent across model scales, with in-domain datasets benefiting the most, but cross-dataset gains remain non-trivial.  
\end{itemize}

These results confirm that combining correctness and visual quality signals yields the most robust multimodal reasoning performance.


\section{Effect of Resolution Perturbation Strength}
\label{appendix:resolution_perturbation_strength}

\begin{table*}[ht!]
\centering
\small
\setlength{\tabcolsep}{3.5pt}
\begin{tabular}{llcccccccccc}
\toprule
& & \multicolumn{10}{c}{\textbf{Model Size}} \\
\cmidrule(lr){3-12}
Train Set & DPO Ratio
& \multicolumn{5}{c}{\textbf{7B}}
& \multicolumn{5}{c}{\textbf{3B}} \\
\cmidrule(lr){3-7} \cmidrule(lr){8-12}
& & HiTab & WikiTQ & VQA & GQA & MathVision
  & HiTab & WikiTQ & VQA & GQA & MathVision \\
\midrule

\multicolumn{12}{l}{\textbf{Train Set: HiTab}} \\
\midrule
& 1.0 vs 0.1
& \cellcolor{bestgreen}\textbf{71.91} & 67.15 & \cellcolor{bestgreen}\textbf{62.65} & \cellcolor{bestgreen}\textbf{42.05} & \cellcolor{bestgreen}\textbf{25.16}
& \cellcolor{bestgreen}\textbf{61.62} & \cellcolor{bestgreen}\textbf{52.15} & \cellcolor{bestgreen}\textbf{61.55} & \cellcolor{bestgreen}\textbf{40.65} & \cellcolor{bestgreen}\textbf{18.78} \\
& 1.0 vs 0.2
& 70.71 & \cellcolor{bestgreen}\textbf{67.95} & 62.40 & 39.90 & 25.07
& 59.72 & 49.70 & 60.40 & 38.15 & 17.20 \\
& 1.0 vs 0.4
& 70.52 & 66.10 & 60.10 & 36.10 & 24.18
& 59.60 & 47.20 & 59.75 & 39.95 & 16.61 \\
& 1.0 vs 0.6
& 67.42 & 65.05 & 61.85 & 38.95 & 24.18
& 58.33 & 46.25 & 59.70 & 39.25 & 16.64 \\
& 1.0 vs 0.8
& 66.79 & 65.45 & 59.90 & 38.50 & 25.13
& 58.96 & 46.60 & 58.05 & 38.70 & 17.40 \\
\midrule

\multicolumn{12}{l}{\textbf{Train Set: VQA}} \\
\midrule
& 1.0 vs 0.1
& \cellcolor{bestgreen}\textbf{68.81} & \cellcolor{bestgreen}\textbf{65.40} & \cellcolor{bestgreen}\textbf{64.80} & \cellcolor{bestgreen}\textbf{44.60} & \cellcolor{bestgreen}\textbf{25.66}
& 60.23 & \cellcolor{bestgreen}\textbf{52.20} & \cellcolor{bestgreen}\textbf{60.70} & \cellcolor{bestgreen}\textbf{41.45} & \cellcolor{bestgreen}\textbf{17.40} \\
& 1.0 vs 0.2
& 68.43 & 63.80 & 64.05 & 44.35 & 25.60
& 59.66 & \cellcolor{bestgreen}\textbf{52.20} & 60.05 & 41.30 & 15.00 \\
& 1.0 vs 0.4
& 65.72 & 63.50 & 60.65 & 41.65 & 25.39
& 59.28 & 51.35 & \cellcolor{bestgreen}\textbf{60.70} & 41.35 & 15.43 \\
& 1.0 vs 0.6
& 66.67 & 61.90 & 60.20 & 40.20 & 25.33
& \cellcolor{bestgreen}\textbf{61.24} & 51.70 & 60.35 & \cellcolor{bestgreen}\textbf{41.45} & 14.47 \\
& 1.0 vs 0.8
& 66.29 & 58.00 & 60.65 & 36.40 & 24.14
& 60.29 & 52.00 & 59.45 & 39.45 & 16.58 \\
\midrule

\multicolumn{12}{l}{\textbf{Train Set: GQA}} \\
\midrule
& 1.0 vs 0.1
& 70.20 & \cellcolor{bestgreen}\textbf{68.20} & \cellcolor{bestgreen}\textbf{63.90} & \cellcolor{bestgreen}\textbf{43.35} & 25.23
& \cellcolor{bestgreen}\textbf{60.23} & \cellcolor{bestgreen}\textbf{52.15} & \cellcolor{bestgreen}\textbf{59.15} & \cellcolor{bestgreen}\textbf{41.25} & \cellcolor{bestgreen}\textbf{17.04} \\
& 1.0 vs 0.2
& 70.20 & 67.00 & 63.10 & 42.10 & 24.26
& 59.66 & 51.70 & 58.70 & 40.05 & 16.68 \\
& 1.0 vs 0.4
& \cellcolor{bestgreen}\textbf{70.77} & 65.10 & 62.90 & 39.60 & 24.34
& 59.34 & 50.95 & 58.15 & 39.40 & 17.14 \\
& 1.0 vs 0.6
& 66.22 & 66.55 & 60.50 & 38.75 & 25.46
& 58.84 & 50.10 & 56.45 & 36.20 & 16.45 \\
& 1.0 vs 0.8
& 66.67 & 67.20 & 59.25 & 38.10 & \cellcolor{bestgreen}\textbf{25.86}
& 59.72 & 48.75 & 58.30 & 38.15 & 15.59 \\
\bottomrule
\end{tabular}
\caption{
Effect of resolution perturbation strength on label-free DPO (VD-LF) across training sets and model sizes.
Each DPO ratio compares the original resolution ($1.0$) with a downsampled counterpart.
Results are reported in accuracy (\%).
For each training set, the best performance per task and model size is highlighted in \cellcolor{bestgreen}green.
}
\label{tab:dpo_ratio_results_horizontal}
\end{table*}

We study how the strength of resolution-induced visual degradation affects the quality of self-generated preference signals in the label-free VD-LF setting.
Although larger resolution gaps can amplify behavioral differences between paired inputs, excessively strong degradation may also corrupt essential visual cues, leading to unstable or noisy supervision.
This experiment aims to identify a practical and robust range of perturbation strengths that yield informative, stable, and scalable preference signals.

\paragraph{Experimental setup.}
For each training dataset (HiTab, VQA, and GQA), we construct preference pairs solely based on resolution perturbations, without using ground-truth answers or correctness filtering.
Each pair consists of a high-resolution image ($1.0$) and a degraded counterpart obtained by downsampling with ratio $\alpha \in \{0.1, 0.2, 0.4, 0.6, 0.8\}$, where smaller $\alpha$ corresponds to more aggressive degradation.
All models are trained under identical VD-LF objectives and optimization settings.
We report results for both Qwen2.5-7B-VL and Qwen2.5-3B-VL to evaluate robustness across model capacities and to assess whether the effect of resolution strength generalizes beyond a single scale.

\paragraph{Results.}
Table~\ref{tab:dpo_ratio_results_horizontal} reveals a clear and highly consistent relationship between resolution perturbation strength and downstream performance across training sets, evaluation benchmarks, and model sizes.
Moderate resolution gaps—most notably $1.0$ vs.\ $0.1$ and $1.0$ vs.\ $0.2$—consistently yield the strongest performance, while both weaker and more aggressive degradations lead to inferior or unstable outcomes.

When trained on HiTab, the 7B model achieves its best overall performance under $1.0$ vs.\ $0.1$, with improvements extending beyond table-centric tasks to general visual reasoning benchmarks such as VQA and GQA.
The same trend holds for the 3B model, where moderate perturbations dominate across nearly all evaluation tasks despite the model’s substantially reduced capacity.
This indicates that resolution-induced preferences provide a strong and capacity-agnostic learning signal.

Similar patterns emerge when training on VQA and GQA.
Across both datasets, moderate perturbations again achieve the best or near-best results on most benchmarks, demonstrating that the observed effect is not tied to a specific training distribution or reasoning format.
In contrast, larger degradation ratios (e.g., $1.0$ vs.\ $0.4$) frequently lead to noticeable drops, particularly on GQA and MathVision, where fine-grained spatial reasoning and visual detail are critical.

Notably, overly mild perturbations ($\alpha \geq 0.6$) consistently underperform despite preserving most visual information.
This suggests that insufficient resolution gaps fail to induce meaningful behavioral divergence between paired inputs, resulting in weak or noisy preference signals.
Overall, the results exhibit a robust unimodal trend with respect to perturbation strength, peaking at moderate resolution gaps.

\paragraph{Discussion.}
These findings highlight resolution perturbation strength as a critical factor in the effectiveness of label-free delta learning.
The observed performance peak at moderate resolution gaps reflects a fundamental trade-off between inducing sufficient behavioral contrast and preserving essential visual semantics.

When the resolution gap is too small, both high- and low-resolution inputs remain visually similar, leading the model to produce nearly indistinguishable responses.
In this regime, the resulting preference signal is weak and provides limited supervisory value beyond noise.
Conversely, when degradation is too severe, key visual cues—such as textual content, table cell boundaries, and spatial relationships—are corrupted to the extent that both responses become unreliable.
This collapses the implicit preference direction and weakens the learning signal.

Moderate perturbations strike a balance between these extremes.
They selectively impair fine-grained visual components while preserving global scene structure, causing the model’s responses to diverge in systematic and semantically meaningful ways.
This exposes consistent multimodal failure modes—such as degraded text recognition or disrupted alignment—that can be effectively exploited by preference learning.
As a result, moderate resolution gaps yield stronger and more stable improvements across tasks.

Importantly, the consistency of this trend across training sets and model sizes suggests that resolution-induced preferences capture intrinsic properties of vision-language robustness rather than dataset-specific artifacts.
Even the smaller 3B model benefits from the same perturbation regime, indicating that the learned signal does not rely on high-capacity reasoning but instead reflects generalizable visual grounding principles.
Taken together, these results position calibrated resolution perturbation as a principled and scalable mechanism for constructing informative label-free preferences in vision-language models.


\section{Generalization to Other Degradations}
\label{appendix:perturbation_comparison}

While VisualDeltas is primarily implemented using \textbf{Resolution Reduction (RD)}---downsampling images to a lower resolution (e.g., 0.1$\times$ original dimensions)---our framework is general and can leverage various types of visual degradation to induce preference pairs. We compare RD against two alternative perturbation strategies: (1) \textbf{Gaussian Noise}, which adds random Gaussian noise to pixel values; and (2) \textbf{Motion Blur}, which applies directional blur to simulate motion artifacts. Both VD-LF and VD-LB use RD as their default perturbation method.

\begin{table*}[ht!]
\centering
\footnotesize
\setlength{\tabcolsep}{4pt}
\renewcommand{\arraystretch}{1.0}
\begin{tabular}{lccccc}
\toprule
Method & HiTab & WikiTQ & VQA & GQA & MathVision \\
\midrule
Baseline & 67.93 & 66.40 & 60.00 & 39.95 & 24.87 \\
\midrule
Gaussian Noise & 72.22 & 67.55 & 63.10 & 43.85 & 25.23 \\
Motion Blur & 72.22 & 68.60 & 64.00 & 42.15 & 24.54 \\
\rowcolor{vdrow}
VD-LF (Ours) & 71.91 & 67.15 & 62.65 & 42.05 & 25.16 \\
\rowcolor{vdrow}
VD-LB (Ours) & 71.40 & 69.90 & 63.55 & 42.90 & 25.13 \\
\bottomrule
\end{tabular}
\caption{
\textbf{Comparison of noise-based methods vs. VisualDeltas on HiTab training set (Qwen2.5-7B-VL).} All values are accuracy (\%). Our methods (VD-LF and VD-LB, both using Resolution Reduction) are highlighted in light blue.
}
\label{tab:noise_vs_vd_comparison_appendix}
\end{table*}

Table~\ref{tab:noise_vs_vd_comparison_appendix} shows that all three perturbation strategies yield comparable improvements across benchmarks. RD (our VD-LF and VD-LB methods) achieves competitive performance on most datasets, while Gaussian Noise and Motion Blur provide similar gains. This suggests that VisualDeltas is not tied to a specific degradation type---the key principle is creating a meaningful quality gap between HQ and LQ views that exposes reasoning brittleness. Importantly, RD is particularly attractive in practice: it is computationally cheap (a single resize operation), deterministic, and works universally across all image types without parameter tuning.

\section{Comparison with Cross-Model Preference Construction}
\label{appendix:large_model_comparison}

This appendix provides detailed results comparing VisualDeltas against cross-model preference construction approaches. We conduct experiments where larger models (32B and 7B) generate preferred responses, while smaller models (7B and 3B) generate dispreferred responses. Table~\ref{tab:large_model_vs_visual_deltas} shows the detailed comparison.

\begin{table*}[!t]
\centering
\scriptsize
\setlength{\tabcolsep}{3pt}
\renewcommand{\arraystretch}{0.95}
\begin{tabular}{llcccccc}
\toprule
\multirow{2}{*}{Paradigm} & \multirow{2}{*}{Method} & \multirow{2}{*}{HiTab} & \multicolumn{4}{c}{Out-of-Domain} \\
\cmidrule(lr){4-7}
& & & WikiTQ & VQA & GQA & MathVision \\
\midrule
\multicolumn{7}{l}{\textbf{Baseline}} \\
& Baseline (no training) & 67.93 & 66.40 & 60.00 & 39.95 & 24.87 \\
\midrule
\multicolumn{7}{l}{\textbf{Large-Model-Guided (External Teacher Required)}} \\
\midrule
& 7B-guided (7B pref vs. 3B dispref)
& 70.01 {\footnotesize(+2.08)}
& 67.25 {\footnotesize(+0.85)}
& 60.90 {\footnotesize(+0.90)}
& 39.45 {\footnotesize(-0.50)}
& \cellcolor{worstred}22.66 {\footnotesize(-2.21)} \\
& 32B-guided (32B pref vs. 7B dispref)
& 67.55 {\footnotesize(-0.38)}
& \cellcolor{bestgreen}\textbf{70.55} {\footnotesize(+4.15)}
& \cellcolor{bestgreen}\textbf{65.10} {\footnotesize(+5.10)}
& \cellcolor{bestgreen}\textbf{43.70} {\footnotesize(+3.75)}
& \cellcolor{worstred}21.88 {\footnotesize(-2.99)} \\
\midrule
\multicolumn{7}{l}{\textbf{Single-Model VisualDeltas (No External Model)}} \\
\midrule
\rowcolor{vdrow}
& VD-LF (HQ vs. LQ)
& \cellcolor{bestgreen}\textbf{71.91} {\footnotesize(+3.98)}
& 67.15 {\footnotesize(+0.75)}
& 62.65 {\footnotesize(+2.65)}
& 42.05 {\footnotesize(+2.10)}
& 25.16 {\footnotesize(+0.29)} \\
\rowcolor{vdrow}
& VD-LB (HQ-correct vs. LQ-wrong)
& 71.40 {\footnotesize(+3.47)}
& \cellcolor{posgreen}69.90 {\footnotesize(+3.50)}
& 63.55 {\footnotesize(+3.55)}
& 42.90 {\footnotesize(+2.95)}
& 25.13 {\footnotesize(+0.26)} \\
\bottomrule
\end{tabular}
\caption{
\textbf{Detailed comparison of large-model-guided preference learning vs. single-model VisualDeltas.} Experiments are conducted on the same 7B target model. \textbf{Large-model-guided} methods require external larger models (32B or 7B) to construct preferences, while \textbf{VisualDeltas} uses only the 7B model itself with resolution-induced deltas. Numbers show absolute accuracy (\%) with gains/losses relative to baseline in parentheses. Best results per column are in bold with dark green background. VisualDeltas rows are highlighted in light blue. Light green indicates positive gains, light red indicates negative changes, and dark red marks the largest drop. Notably, VD-LF achieves the best HiTab performance (71.91\%), and VisualDeltas methods maintain positive gains across all benchmarks unlike large-model-guided methods which hurt MathVision.
}
\label{tab:large_model_vs_visual_deltas}
\end{table*}

\paragraph{Key Findings:}
\textbf{Better In-Domain Performance:} On HiTab, VD-LF (71.91\%) outperforms the 32B-guided baseline (67.55\%). This suggests that resolution-induced self-correction is more effective for visual robustness than learning from a larger model that may have different visual biases.

\textbf{Comparable Generalization:} On out-of-domain tasks like WikiTQ, VD-LB achieves gains (+3.50\%) comparable to 32B-guided training (+4.15\%), but avoids the substantial computational cost and latency of querying a larger model.

\textbf{Avoiding Cross-Model Failures:} Note that 32B-guided training harms MathVision performance (-2.99\%), likely due to distribution shift. VisualDeltas maintains positive gains (+0.26\%), demonstrating superior stability.

\end{document}